\documentclass[runningheads]{llncs}

\usepackage{eccv}
\usepackage{eccvabbrv}
\usepackage{graphicx}
\usepackage{booktabs}
\usepackage{multirow}
\usepackage[ruled,vlined,linesnumbered]{algorithm2e}
\SetAlFnt{\small}
\SetAlgoNlRelativeSize{-1}
\DontPrintSemicolon

\usepackage[accsupp]{axessibility}
\usepackage{hyperref}
\begin{document}

\title{ConTrack: Constrained Hand Motion Tracking with Adaptive Trade-off Control}
\titlerunning{ConTrack}
\author{Yutong Liang \and Quanquan Peng \and Ri-Zhao Qiu \and Xiaolong Wang}
\authorrunning{Y. Liang et al.}
\institute{University of California San Diego, La Jolla, CA, USA}

\maketitle

\begin{abstract}
    Human demonstrations provide strong priors for robot manipulation, yet it is non-trivial to transfer them to execute on real robots due to the kinematic gap. In dexterous manipulation, it remains challenging to track long-horizon, contact-rich sequences even in simulators: a reference-tracking policy must keep objects on their target trajectories while preserving demonstrated joint motion and contact timing. Existing approaches often rely on hand-crafted reward tuning that require per-sequence tuning and break under limited interaction budgets. We introduce ConTrack, a reinforcement learning (RL) framework that scales with tracking data. ConTrack treats object tracking as a constraint and allocates remaining control authority to motion fidelity, which allows it to adapt task--style trade-offs online using a dual-variable update. In addition, ConTrack also stabilizes long-horizon learning with an adaptive mid-trajectory reset library that reuses policy-reachable simulator states. Our qualitative and quantitative results in simulation tracking and real robot demonstrate that ConTrack improves success and object pose accuracy significantly over prior arts while preserving joint and contact fidelity. Website: \url{https://www.lyt0112.com/projects/ConTrack}.

\end{abstract}

\section{Introduction}
\label{sec:introduction}

Recent progress in robot manipulation has been driven by scaling robot data and training general-purpose policies~\cite{barreiros2025-lbm,intelligence2025pi_05,liu2024rdt,spiritai2026-spiritv15,peng2026fact}. A parallel trend aims to learn from humans, motivated by the abundance of human demonstrations and their potential to supervise dexterous behavior at scale~\cite{cai2025-in-n-on,qiu2025-humanoid,kareer2025-emergence,kareer2025-egomimic,luo2026-being,wen2025-grdexter,zheng2026egoscale, luo2025sonic}. Yet many of these pipelines extract high-level intent or coarse motion cues from human observations and still rely on an additional stage to calibrate end-effector interaction for contact-rich manipulation~\cite{cai2025-in-n-on,zheng2026egoscale}.

\begin{figure}[t]
  \centering
  \includegraphics[width=\linewidth]{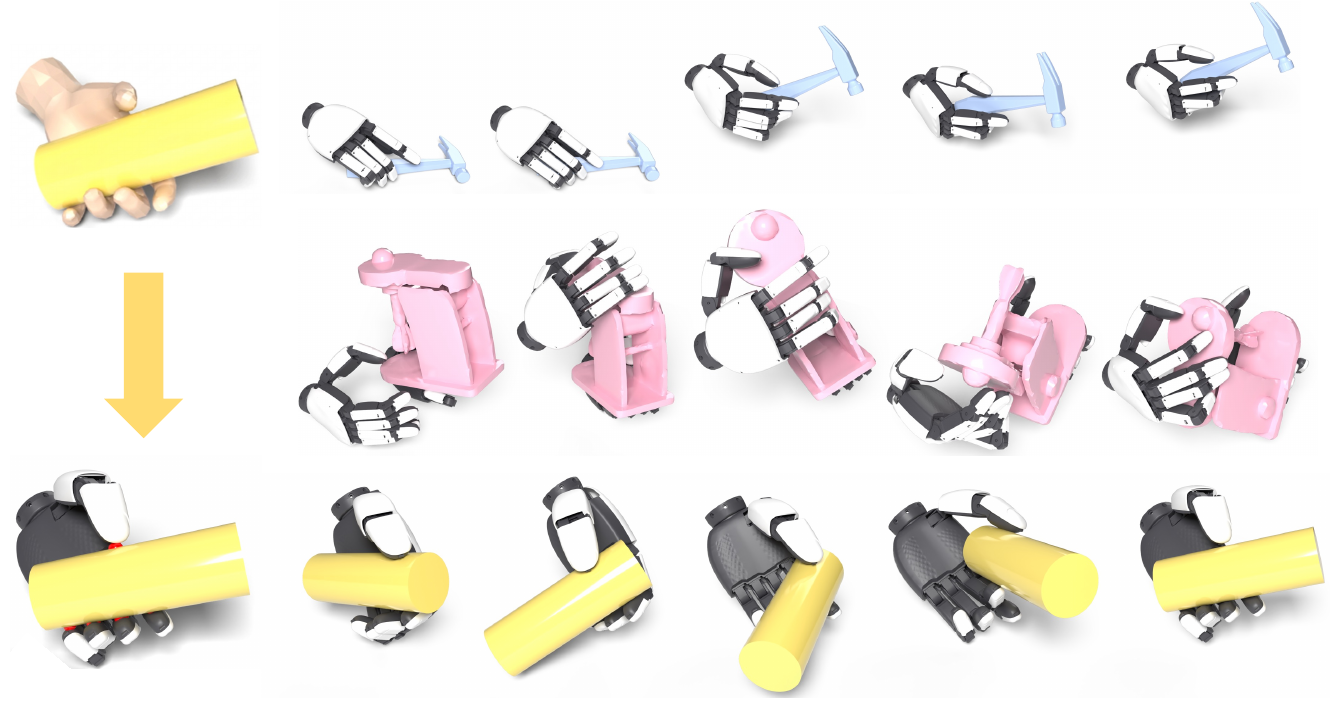}
  \caption{ConTrack enables long-horizon, contact-rich dexterous hand tracking. With adaptive task--style trade-off control, ConTrack yields physically plausible object motion balancing hand-object contacts and geometric tracking. This aligns the embodiment gap between human hands and dexterous hands while transferring human dexterous play data to real robots (Tasks from top to bottom: functional tool usage, articulated object interaction, and in-hand rotation).}
  \label{fig:teaser}
\end{figure}

This extra stage matters because dexterous manipulation is dominated by contact, and contact is where human-to-robot transfer is least forgiving. Compared to collecting robot-native demonstrations through teleoperation interfaces~\cite{cheng2024-opentv,chi2024-umi,xu2025dexumi,liang2026fingr}, human data introduces a systematic mismatch between the demonstrated hand motion and what a robot hand can physically realize. The root cause is morphological and actuation differences between human hands and robot end effectors, which manifest as a kinematic gap when one attempts to execute human motion directly.

One intuitive way to address the kinematic gaps is retargeting. Optimization-based retargeting aligns a human hand trajectory to a robot hand by minimizing geometric errors, supported by rigid-body dynamics toolkits~\cite{carpentier2019-pinocchio} and vision-based teleoperation systems that infer targets~\cite{qin2023-anyteleop,handa2020-dexpilot}. Neural retargeting further improves scalability by learning fast mappings from human to robot kinematics~\cite{yin2025-geometric-retargeting}. Purely geometric objectives tend to be fragile once contact schedules matter. Recent work incorporates physical feasibility and object-centric goals, either through physics-informed retargeting that reasons about contact, or through functional retargeting that prioritizes object motion over joint-level imitation~\cite{pan2025-spider,mandi2025-dexmachina}. Learned tracking controllers further reduce per-clip optimization by training neural policies to follow time-indexed references~\cite{liu2025-dextrack}. These approaches still leave an open question: when the reference is not jointly realizable, how should a system choose a trade-off between task success and motion style without per-sequence manual tuning.

In other domains, tracking-centric formulations have reduced reliance on manual reward engineering and improved scalability with diverse motion data~\cite{peng2018deepmimic,peng2022ase,tessler2024maskedmimic,liao2025beyondmimic,ze2025twist,luo2025sonic}. Inspired by this shift, we propose \textbf{ConTrack}, a reference-tracking reinforcement learning framework for contact-rich hand--object interaction in physics simulation. ConTrack introduces an online task--style mixing mechanism that adapts the trade-off during training via a dual-variable update, and an adaptive mid-trajectory reset library that stabilizes long-horizon learning by restarting from policy-reachable simulator states near the current failure boundary. The resulting policy can deviate from the reference when required by physics while remaining anchored to the demonstrated motion and contact cues.

We present quantitative and qualitative evaluations on GRAB~\cite{taheri2020grab}, ARCTIC~\cite{fan2023arctic}, and DexterHand~\cite{liang2026dextercap}. ConTrack improves success and object pose accuracy while preserving joint and contact fidelity. We also include a real-world feasibility study on a bimanual xArm7+xHand platform to verify that the learned trajectories are executable on real hardware.

In sum, our contributions are:

\begin{itemize}
\item An online adaptive task--style mixing mechanism based on a dual-variable update.
\item An adaptive mid-trajectory reset library that reuses policy-reachable simulator states.
\item A framework that transforms physically-infeasible reference motion into executable robotic trajectories, evaluated across three benchmark tiers of increasing contact complexity and validated on real-world bimanual manipulation setup.
\end{itemize}

\section{Related Work}
\label{sec:related_work}

\subsubsection{Dexterous manipulation from human demonstrations}
Human hand--object recordings are used as supervision for robotic dexterity, with simulation and policy learning bridging the embodiment gap from captured kinematics to robot actuation~\cite{taheri2020grab,fan2023arctic,liang2026dextercap,brahmbhatt2020contactpose,chao2021dex}. The supervision signal spans larger motion-capture datasets, touch sensing, and video or teleoperation sources~\cite{jian2023affordpose,jiang2023chairs,fu2025gigahands,lu2025humoto,kim2024parahome,song2025-opentouch,hsieh2025dexman,zhao2026dexh2r,cheng2024-opentv,chi2024-umi,xu2025dexumi,jiang2025gsworld,qin2023-anyteleop,jiang2026crosshand,geng2025roboverse,handa2020-dexpilot}. To make references executable, methods range from object-centric conditioning~\cite{chen2024-object} and residual reinforcement learning on top of imitation~\cite{li2025-maniptrans} to tracking-centric controllers~\cite{liu2025-dextrack} and functional retargeting that prioritizes object outcomes~\cite{mandi2025-dexmachina}; across settings, task success and motion fidelity can disagree, so a learner must allocate limited interaction budget to resolve the trade-off.

\subsubsection{Reference tracking in physics simulation}
Reference tracking has a long history in physics-based character control, where motion capture clips specify the desired behavior and reinforcement learning discovers feedback control that is robust to perturbations and missing information~\cite{peng2018deepmimic,peng2022ase,tessler2024maskedmimic,ze2025twist,liao2025beyondmimic,luo2025sonic}. Long-horizon training often benefits from curricula that emphasize later states and from exploration strategies that expand the reachable set of states over time~\cite{resnick2018-backplay,ecoffet2019-goexplore,xu2025dexplore}. These ideas motivate mid-trajectory training, but dexterous hand--object interaction introduces a sharp complication. Directly resetting to the reference can be physically inconsistent because contacts and object configurations are coupled through discontinuous dynamics. This mismatch motivates ConTrack's emphasis on policy-reachable mid-trajectory states, which preserves the efficiency benefits of mid-clip training while avoiding unrecoverable starts induced by infeasible reference configurations.

\subsubsection{Constrained and multi-objective reinforcement learning}
Balancing task tracking against imitation is often implemented through a single shaped reward that mixes task terms and style terms with fixed weights~\cite{peng2018deepmimic,peng2022ase,tessler2024maskedmimic,li2025-maniptrans,liu2025-dextrack}. In clip-conditioned tracking with contact discontinuities, fixed mixtures can require retuning as task difficulty and signal scale shift across clips and across phases of a clip~\cite{li2025-maniptrans,liu2025-dextrack,mandi2025-dexmachina}. Constrained reinforcement learning instead treats task success as a requirement and optimizes a secondary objective within the feasible set, commonly via primal--dual updates, reward-constrained policy optimization, and Lagrangian relaxations~\cite{achiam2017-cpo,tessler2018-rcpo,ishihara2025-constraints}. Closely related work studies how to learn stylistic behavior from imperfect demonstrations under a task optimality constraint~\cite{wen2025-constrained}. ConTrack adopts this constrained viewpoint, using the normalized task constraint and an online dual update to adjust the task--style allocation during training.

\section{Method}
\label{sec:method}

ConTrack models object tracking problem as a constraint and allocates remaining optimization pressure to style fidelity. Three components make this practical under a fixed interaction budget: an online dual controller (Sec.~\ref{sec:adaptive_mixing}) that adjusts the task--style allocation, a reset library (Sec.~\ref{sec:reset_library}) that restarts rollouts from policy-reachable states near the current failure boundary, and contact priors (Sec.~\ref{sec:contact_priors}) that anchor the style objective to reference interaction patterns.

\subsection{Preliminaries}
\label{sec:problem_setup}
In our setup, each reference clip induces a finite-horizon MDP with horizon $T$ and discrete time index $t\in\{0,\dots,T-1\}$. The simulator has physical state $x_t\in\mathcal{X}$, including robot joints, object poses, and their velocities. The reference provides time-indexed targets. It includes a joint trajectory $q^{\mathrm{ref}}_t\in\mathbb{R}^D$ over $D$ actuated joints and object pose targets for $O$ objects, given by translations $p^{\mathrm{ref}}_{t,o}\in\mathbb{R}^3$ and unit quaternions $\bar{q}^{\mathrm{ref}}_{t,o}\in\mathbb{S}^3$ for $o\in\{1,\dots,O\}$. Additionally, we provide link-level contact annotations for $L$ hand links, given by binary contact events $c^{\mathrm{ref}}_{t,o,\ell}\in\{0,1\}$ and object-local contact points $y^{\mathrm{ref}}_{t,o,\ell}\in\mathbb{R}^3$ for $\ell\in\{1,\dots,L\}$. The joint reference $q^{\mathrm{ref}}_{0:T-1}$ is obtained by retargeting captured human hand motion to the robot embodiment~\cite{carpentier2019-pinocchio,yin2025-geometric-retargeting,pan2025-spider}.

We treat the reference index as part of the state and define $s_t=(x_t,t)$. A stochastic policy $\pi(a_t\mid s_t)$ selects an action $a_t\in\mathbb{R}^D$. The agent receives a reward that we decompose as
\begin{equation}
  r(s_t,a_t)=r_g(s_t,a_t)+r_s(s_t,a_t)+r_p(s_t,a_t),
\end{equation}
where $r_g$ measures task success through object pose tracking, $r_s$ measures style fidelity through hand kinematics and contacts, and $r_p$ penalizes high-frequency motion.

The policy controls the robot through residual joint targets around the reference. At frame $t$, the policy outputs a residual joint displacement $a_t$ and we set the joint position target as: $q^{\mathrm{tar}}_t = q^{\mathrm{ref}}_t + a_t$.

This parameterization keeps exploration near the reference while allowing corrective deviations that compensate for kinematic and dynamic mismatch.

We evaluate task and style with discounted returns using a discount factor $\gamma\in(0,1]$
\begin{equation}
  J_g(\pi)=\mathbb{E}_\pi\left[\sum_{t=0}^{T-1}\gamma^t r_g(s_t,a_t)\right],\qquad
  J_s(\pi)=\mathbb{E}_\pi\left[\sum_{t=0}^{T-1}\gamma^t r_s(s_t,a_t)\right].
\end{equation}
ConTrack maximizes style while meeting a minimum level of task success:
\begin{equation}
  \max_{\pi}\; J_s(\pi)\quad \text{s.t.}\quad J_g(\pi)\ge \alpha J_g^\star,
  \label{eq:cmdp_form}
\end{equation}
where $\alpha\in(0,1]$ is a target task ratio and $J_g^\star$ is a clip-specific normalizer defined as a running maximum of a moving estimate of $J_g$ during training. The ratio $J_g/J_g^\star$ is dimensionless and makes the constraint comparable across clips with different task scales.

\subsection{Adaptive Task--Style Mixing}
\label{sec:adaptive_mixing}
Fixed reward weights are brittle in contact-rich tracking because difficulty varies across clips and within a single clip. ConTrack therefore adapts the task--style trade-off online using a scalar controller driven by the normalized task return.

We connect Eq.~\ref{eq:cmdp_form} to an online weight controller through a Lagrangian relaxation~\cite{achiam2017-cpo,tessler2018-rcpo,wen2025-constrained}
\begin{equation}
  L(\pi,\lambda)=J_s(\pi)-\lambda\left(\alpha-\frac{J_g(\pi)}{J_g^\star}\right),
  \label{eq:lagrangian}
\end{equation}
where $\lambda\in\mathbb{R}$ is a scalar controller state. Larger $\lambda$ increases pressure to satisfy the task constraint, while smaller $\lambda$ allocates more optimization capacity to style.

PPO~\cite{schulman2017-ppo} uses an advantage estimate to update the policy. We compute three advantage estimates, $A_g$ from the task reward $r_g$, $A_s$ from the style reward $r_s$, and $A_p$ from the penalty reward $r_p$, and update the policy using a mixed advantage
\begin{equation}
  A_{\mathrm{mix}} = w_{\mathrm{task}} A_g + (1-w_{\mathrm{task}}) A_s + A_p,
\end{equation}
where $w_{\mathrm{task}}\in[0,1]$ is shared across all parallel environments of the clip. We map this scalar state to a convex mixing weight
\begin{equation}
  w_{\mathrm{task}} = \sigma(\lambda),
\end{equation}
where $\sigma$ is the logistic sigmoid.

We update $\lambda$ from an online estimate of the task return $J_g$. Let $\hat{J}_g$ be a running estimate of $J_g$ under the current policy, and let $J_g^\star$ be its running maximum. The update is
\begin{equation}
  \lambda \leftarrow \lambda + \eta\left(\alpha - \frac{\hat{J}_g}{J_g^\star}\right),
  \label{eq:dual_update}
\end{equation}
where $\eta>0$ is a step size. The ratio $\hat{J}_g/J_g^\star$ is scale-free across clips. When it drops below $\alpha$, Eq.~\ref{eq:dual_update} increases $w_{\mathrm{task}}$ and shifts optimization toward task tracking. When it stays above $\alpha$, the update decreases $w_{\mathrm{task}}$ and allocates more capacity to style.

\begin{algorithm}[t]
  \caption{CMDP-style task constraint with an online dual controller}
  \label{alg:cmdp}
  \KwIn{target ratio $\alpha$, step size $\eta$}
  \KwData{dual state $\lambda$, running task estimate $\hat{J}_g$, running maximum $J_g^\star$}
  \While{training}{
    collect rollouts under the current reset distribution\tcp*[r]{$\triangleright$ Sec.~\ref{sec:reset_library}}
    $G_g \leftarrow$ discounted episodic return from $r_g$\;
    $\hat{J}_g \leftarrow \mathrm{runningmean}(G_g)$;\quad $J_g^\star \leftarrow \max\{J_g^\star,\hat{J}_g\}$\;
    $\lambda \leftarrow \lambda + \eta\left(\alpha-\hat{J}_g/J_g^\star\right)$\tcp*[r]{$\triangleright$ Eq.~\ref{eq:dual_update}}
    $w_{\mathrm{task}} \leftarrow \sigma(\lambda)$\;
    compute $A_g,A_s,A_p$\;
    $A_{\mathrm{mix}} \leftarrow w_{\mathrm{task}}A_g+(1-w_{\mathrm{task}})A_s+A_p$\;
    update $\pi$ using $A_{\mathrm{mix}}$\;
  }
\end{algorithm}

\subsection{Adaptive Mid-Trajectory Reset Library}
\label{sec:reset_library}
Long-horizon tracking is inefficient when training always starts from the first reference frame, and curriculum strategies that restart from later states can substantially improve coverage~\cite{resnick2018-backplay,ecoffet2019-goexplore}. Early failures dominate the rollout distribution, and later contact phases receive few learning updates. ConTrack addresses this with a mid-trajectory reset library indexed by reference frame. For each frame $k\in\{0,\dots,T-1\}$, the library stores a simulator state that contains robot joints and velocities and object poses and velocities. Resetting to a stored state initializes the simulator consistently and sets the current reference index to $k$.

\paragraph{Online refresh.}
In human motion tracking without object interaction, resets are often initialized by setting the simulator directly to the reference motion at a chosen frame~\cite{peng2018deepmimic,peng2022ase,tessler2024maskedmimic}. In hand--object interaction, such resets can be physically inconsistent because contact configurations may not be realizable under the simulator dynamics. Resetting by copying the reference state can therefore produce unrecoverable starts. ConTrack instead refreshes the reset library from empirically reachable states collected from rollouts of the current policy. Consider a completed episode that starts at frame $t_{\mathrm{start}}$ and terminates at frame $t_{\mathrm{end}}$. For each visited frame $k\in[t_{\mathrm{start}},t_{\mathrm{end}}]$ we define the continuation length
\begin{equation}
  \ell_k=t_{\mathrm{end}}-k+1.
\end{equation}
We maintain an exponential moving average $\bar{\ell}_k$ of $\ell_k$ and a running maximum $\ell_k^{\max}$. If a rollout achieves $\ell_k>\ell_k^{\max}$, we overwrite the stored state at frame $k$ with the visited simulator state and set $\ell_k^{\max}=\ell_k$. This keeps the library aligned with policy-reachable states that support longer continuations.

\paragraph{Difficulty-aware sampling.}
We define a survival ratio
\begin{equation}
  u_k=\frac{\bar{\ell}_k}{T-k},
  \label{eq:survival_ratio}
\end{equation}
and sample reset frames with
\begin{equation}
  p(k)\propto \exp\left(-\frac{u_k}{\tau}\right),
  \label{eq:reset_sampling}
\end{equation}
where $\tau>0$ is a temperature parameter. This concentrates training on segments where the policy tends to fail quickly relative to the remaining horizon. This produces rollouts that repeatedly practice entering difficult contact phases, and the reset distribution moves earlier as tracking improves.

\begin{algorithm}[t]
  \caption{Adaptive mid-trajectory reset library}
  \label{alg:reset_library}
  \KwIn{clip length $T$, temperature $\tau$, smoothing coefficient $\beta$}
  \KwData{for all k$\in [0, T-1]$: cached simulator states $B[k]$, EMA continuation length $\bar{\ell}_k$, best length $\ell_k^{\max}$}
  \ForEach{completed episode}{
    $t_s \leftarrow$ start frame;\quad $t_e \leftarrow$ end frame\;
    \For{$k \leftarrow t_s$ \KwTo $t_e$}{
      $\ell \leftarrow t_e-k+1$;\quad $\bar{\ell}_k \leftarrow (1-\beta)\bar{\ell}_k + \beta \ell$\;
      \If{$\ell > \ell_k^{\max}$}{
        $B[k] \leftarrow$ simulator state at frame $k$;\; $\ell_k^{\max} \leftarrow \ell$
      }
    }
  }
  $u_k \leftarrow \bar{\ell}_k/(T-k)$ for all $k$\tcp*[r]{$\triangleright$ survival ratio, Eq.~\ref{eq:survival_ratio}}
  sample $k \propto \exp\left(-u_k/\tau\right)$\tcp*[r]{$\triangleright$ Eq.~\ref{eq:reset_sampling}}\;
  reset simulator to $B[k]$ and set reference index to $k$\;
\end{algorithm}

\subsection{Contact Priors}
\label{sec:contact_priors}
We use reference contact annotations as an auxiliary style prior. Let $c_{t,o,\ell}\in\{0,1\}$ denote the contact event between link $\ell$ and object $o$ observed in simulation at frame $t$, and let $y_{t,o,\ell}\in\mathbb{R}^3$ denote the corresponding object-local contact point. The style reward includes a contact event overlap term that rewards agreement with the reference contact events $c^{\mathrm{ref}}_{t,o,\ell}$. When both the simulator and reference indicate contact, we additionally penalize the distance between $y_{t,o,\ell}$ and the annotated contact point $y^{\mathrm{ref}}_{t,o,\ell}$.

\begin{figure}[t]
  \centering
  \includegraphics[width=\linewidth]{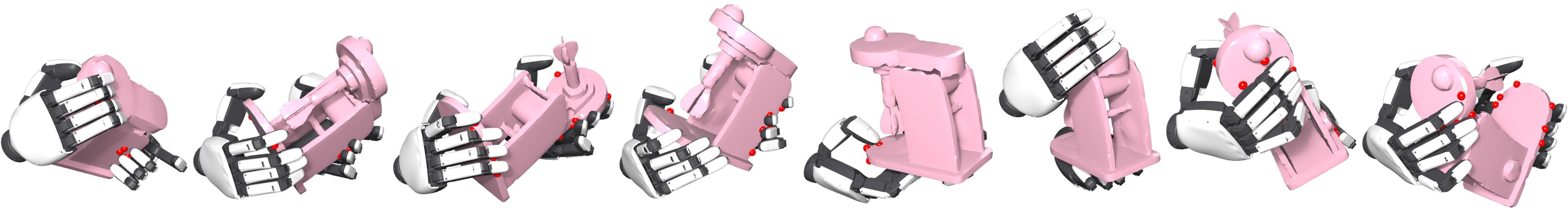}
  \caption{Contact priors from data. Red dots mark reference object contact points. During reinforcement learning, the style objective encourages each contacting link to match its target contact.}
  \label{fig:contact_prior_vis}
\end{figure}

We map dataset contact annotations to the robot link set and use them as link-level contact targets. Fig.~\ref{fig:contact_prior_vis} visualizes the targets. We evaluate contact fidelity with explicit event and point metrics and ablate contact priors in Sec.~\ref{sec:experiments}. See appendix for annotation processing.

\section{Experiments}
\label{sec:experiments}

\subsection{Benchmarks}
\label{sec:benchmarks}

\begin{figure}[t]
  \centering
  \begin{subfigure}{0.49\linewidth}
    \centering
    \includegraphics[width=\linewidth]{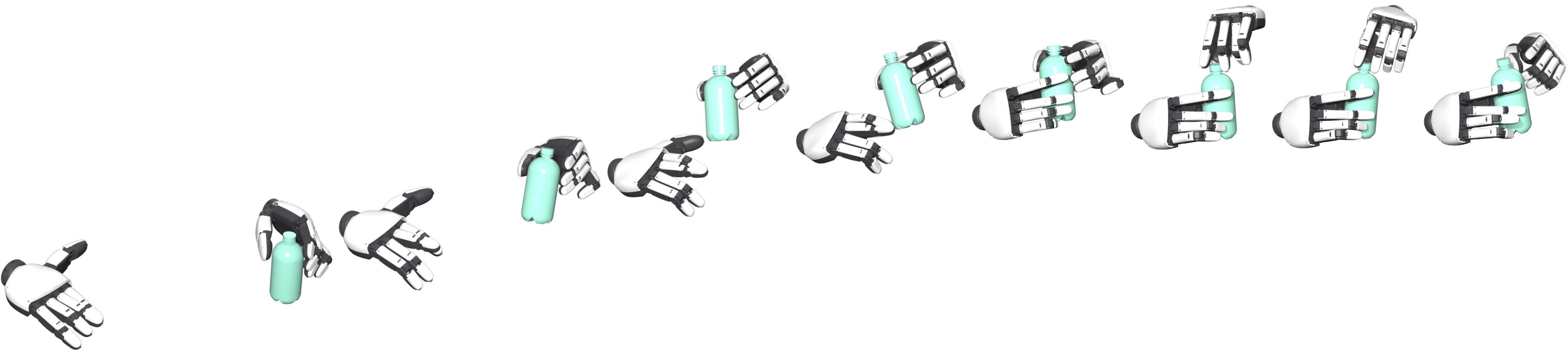}
    \caption{\texttt{GRAB Waterbottle Offhand}}
  \end{subfigure}
  \hfill
  \begin{subfigure}{0.49\linewidth}
    \centering
    \includegraphics[width=\linewidth]{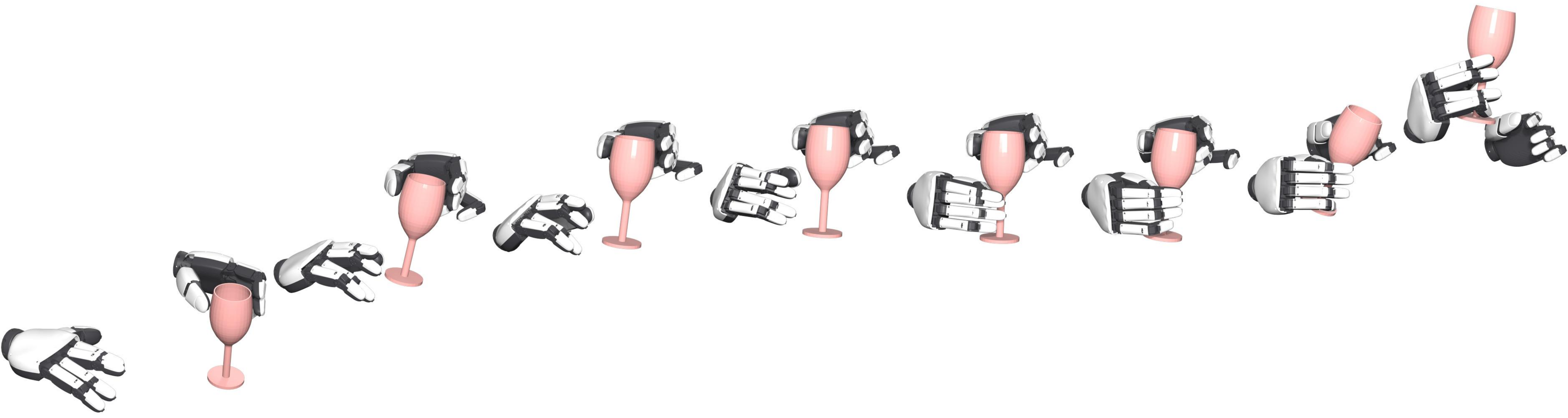}
    \caption{\texttt{GRAB Wineglass Offhand}}
  \end{subfigure}

  \begin{subfigure}{0.49\linewidth}
    \centering
    \includegraphics[width=\linewidth]{figures/sim_arctic_s01_waffleiron_use_01.pdf}
    \caption{\texttt{ARCTIC Waffleiron Use}}
  \end{subfigure}
  \hfill
  \begin{subfigure}{0.49\linewidth}
    \centering
    \includegraphics[width=\linewidth]{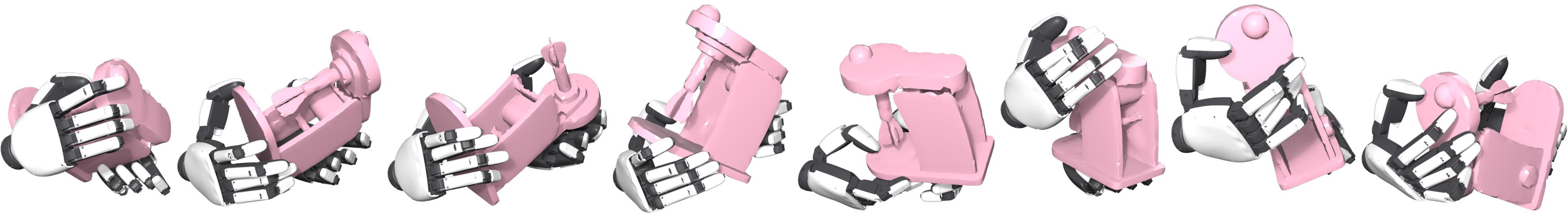}
    \caption{\texttt{ARCTIC Mixer Use}}
  \end{subfigure}

  \begin{subfigure}{0.49\linewidth}
    \centering
    \includegraphics[width=\linewidth]{figures/sim_arctic_s01_box_use_01.pdf}
    \caption{\texttt{ARCTIC Box Use}}
  \end{subfigure}
  \hfill
  \begin{subfigure}{0.49\linewidth}
    \centering
    \includegraphics[width=\linewidth]{figures/sim_dexterhand_ring.pdf}
    \caption{\texttt{DexterHand Ring}}
  \end{subfigure}

  \begin{subfigure}{0.49\linewidth}
    \centering
    \includegraphics[width=\linewidth]{figures/sim_dexterhand_cuboid.pdf}
    \caption{\texttt{DexterHand Cuboid-0}}
  \end{subfigure}
  \hfill
  \begin{subfigure}{0.49\linewidth}
    \centering
    \includegraphics[width=\linewidth]{figures/sim_dexterhand_cuboid_01.pdf}
    \caption{\texttt{DexterHand Cuboid-1}}
  \end{subfigure}
  \caption{Simulation snapshots.}
  \label{fig:sim_snapshots}
\end{figure}
\begin{table}[t]
  \caption{Benchmark clips and reference-only statistics. Max disp: max object AABB diagonal displacement. End2End rot: end-to-end rotation between first and last frame. Contact occ.: fraction of frames with any hand--object contact.}
  \label{tab:benchmark_stats}
  \centering
  \scriptsize
  \resizebox{\linewidth}{!}{
  \begin{tabular}{@{}cccccccc@{}}
    \toprule
    Dataset & Clip & Time(s) & Hand & Articulated & Max disp(m) & End2End rot(rad) & Contact occ. \\
    \midrule
    \multirow{5}{*}{GRAB} & \texttt{cubemedium\_offhand} & 3.0 & bimanual & False & 0.581 & 1.426 & 0.819 \\
     & \texttt{cylindermedium\_offhand} & 3.0 & bimanual & False & 0.805 & 2.675 & 0.942 \\
     & \texttt{hammer\_use} & 6.0 & bimanual & False & 0.304 & 0.001 & 0.878 \\
     & \texttt{waterbottle\_offhand} & 4.0 & bimanual & False & 0.470 & 0.439 & 0.808 \\
     & \texttt{wineglass\_offhand} & 3.0 & bimanual & False & 0.595 & 0.751 & 0.967 \\
    \midrule
    \multirow{4}{*}{ARCTIC} & \texttt{box\_use} & 4.1 & bimanual & True & 0.140 & 0.171 & 1.000 \\
     & \texttt{mixer\_use} & 10.0 & bimanual & True & 0.271 & 1.785 & 1.000 \\
     & \texttt{notebook\_use} & 3.0 & bimanual & True & 0.272 & 2.962 & 1.000 \\
     & \texttt{waffleiron\_use} & 4.0 & bimanual & True & 0.231 & 0.358 & 1.000 \\
    \midrule
    \multirow{4}{*}{DexterHand} & \texttt{Cuboid\_00} & 3.0 & single-hand & False & 0.098 & 3.082 & 1.000 \\
     & \texttt{Cuboid\_01} & 2.0 & single-hand & False & 0.060 & 2.849 & 1.000 \\
     & \texttt{Cylinder} & 5.0 & single-hand & False & 0.113 & 2.477 & 1.000 \\
     & \texttt{Ring} & 5.0 & single-hand & False & 0.105 & 2.415 & 1.000 \\
    \bottomrule
  \end{tabular}}
\end{table}

We evaluate ConTrack on reference clips spanning bimanual interaction and dexterous in-hand manipulation. Each clip defines an independent reference-tracking task with horizon $T$, and we train one policy per clip. We group clips into three tiers of increasing contact complexity. \textbf{GRAB}~\cite{taheri2020grab} focuses on bimanual rigid-object interaction. \textbf{ARCTIC}~\cite{fan2023arctic} adds articulated objects and multi-object contact. \textbf{DexterHand}~\cite{liang2026dextercap} targets continuous single-hand in-hand rotation. Table~\ref{tab:benchmark_stats} summarizes clip statistics computed from the reference signals. Fig.~\ref{fig:sim_snapshots} shows representative simulated rollouts, and additional rollouts are shown in appendix.

\subsection{Training and Evaluation Protocol}
All learning-based methods are trained for 5000 PPO~\cite{schulman2017-ppo} updates per clip under a fixed simulator-step budget. Mid-trajectory resets redistribute interaction along the clip without increasing total steps. During training each method uses its own reset strategy. For evaluation we always reset from the first reference frame so that progress reflects end-to-end tracking.

We evaluate with fixed pose-break thresholds of $0.10$\,m translation error and $1.00$\,rad rotation error. Let $p_{t,o}\in\mathbb{R}^3$ and $\bar{q}_{t,o}\in\mathbb{S}^3$ denote the simulated translation and unit quaternion of object $o$ at frame $t$, and let $p^{\mathrm{ref}}_{t,o}$ and $\bar{q}^{\mathrm{ref}}_{t,o}$ be the reference targets. Define $\Delta p_{t,o}=\lVert p_{t,o}-p^{\mathrm{ref}}_{t,o}\rVert_2$ and $\Delta \theta_{t,o}=2\arccos\left(\left|\bar{q}_{t,o}^\top \bar{q}^{\mathrm{ref}}_{t,o}\right|\right)$. An episode terminates at the first index $\tau$ such that some object violates $\Delta p_{\tau,o}>0.10$ or $\Delta \theta_{\tau,o}>1.00$, or at $\tau=T-1$ if it reaches the end. We report success rate SR as the fraction of evaluation episodes with $\tau=T-1$, and \emph{progress} as the normalized termination index
$\mathrm{progress}=\frac{\tau}{T-1}$,
where $\tau$ is the termination frame index.

\subsection{Metrics}
We report both task tracking and style fidelity. Task metrics are mean object translation error in meters and mean object rotation error in radians with respect to the reference object tracks. For an episode with termination index $\tau$, we compute
\begin{equation}
  E_{\mathrm{pos}}=\frac{1}{(\tau+1)O}\sum_{t=0}^{\tau}\sum_{o=1}^{O}\Delta p_{t,o},\qquad
  E_{\mathrm{rot}}=\frac{1}{(\tau+1)O}\sum_{t=0}^{\tau}\sum_{o=1}^{O}\Delta \theta_{t,o}.
\end{equation}
Style metrics include mean arm and finger joint position errors in radians with respect to the reference joint trajectories. Let $q_t\in\mathbb{R}^D$ denote the robot joint configuration and $q^{\mathrm{ref}}_t$ the reference target at frame $t$. We report the mean absolute joint error, aggregated over frames up to $\tau$, and separately averaged over arm and finger joints.

To quantify contact fidelity, we aggregate contact event classification over time and over annotated link--object pairs. Let $c_{t,o,\ell}\in\{0,1\}$ denote whether link $\ell$ contacts object $o$ at frame $t$ in simulation, and let $c^{\mathrm{ref}}_{t,o,\ell}$ denote the reference contact annotation. We compute precision, recall, and F1 from the aggregated confusion matrix of $c_{t,o,\ell}$ against $c^{\mathrm{ref}}_{t,o,\ell}$. We also report contact point error in meters in object-local coordinates on matched contact events. Since early termination shortens the effective horizon, we always report progress alongside error metrics.

\subsection{Baselines and Ablations}
We compare ConTrack against three baselines that resolve the task--style conflict in different ways: ManipTrans~\cite{li2025-maniptrans}\footnote{We reimplement ManipTrans to match our environment and evaluation setup.}, DexMachina~\cite{mandi2025-dexmachina}\footnote{DexMachina's code is not fully open source, so we integrated their Virtual Object Controllers into our codebase.}, and SPIDER~\cite{pan2025-spider}. ManipTrans starts from an imitative controller and uses residual reinforcement learning to correct execution errors under contact dynamics. DexMachina targets functional retargeting with object-centric goals. SPIDER constructs a physically feasible retargeted trajectory without policy learning. We evaluate all methods under the same termination rule and metrics.

We study three ablations that isolate the mechanisms responsible for stability and trade-off control. The first ablation replaces adaptive task--style mixing in Sec.~\ref{sec:adaptive_mixing} with fixed mixing weights. The second ablation replaces the reset library in Sec.~\ref{sec:reset_library} with simpler reset strategies that do not adapt to the current failure boundary. The third ablation removes contact priors from the style objective. Unless stated otherwise, all variants use the same observations, action space, and termination rule.

\subsection{Main Results}
\label{sec:results}

\begin{table}[t]
  \caption{Main experiment: baseline comparison under the fixed 5000-update budget. We report mean and standard deviation across clips.}
  \label{tab:baselines}
  \centering
  \scriptsize
  \resizebox{\linewidth}{!}{
    \begin{tabular}{@{}ccccccc@{}}
      \toprule
      Method     & Progress $\uparrow$        & Obj pos (m) $\downarrow$   & Obj rot (rad) $\downarrow$ & Finger err (rad) $\downarrow$ & Contact F1 $\uparrow$      & Contact pt (m) $\downarrow$ \\
      \midrule
      Ours       & \textbf{0.899} $\pm$ 0.195 & 0.026 $\pm$ 0.006          & 0.272 $\pm$ 0.105          & 0.163 $\pm$ 0.014             & \textbf{0.784} $\pm$ 0.072 & \textbf{0.018} $\pm$ 0.005  \\
      ManipTrans & 0.743 $\pm$ 0.292          & \textbf{0.012} $\pm$ 0.009 & \textbf{0.207} $\pm$ 0.078 & 0.277 $\pm$ 0.089             & 0.620 $\pm$ 0.068          & 0.030 $\pm$ 0.010           \\
      DexMachina & 0.246 $\pm$ 0.052          & 0.038 $\pm$ 0.018          & 0.348 $\pm$ 0.121          & \textbf{0.147} $\pm$ 0.016    & 0.708 $\pm$ 0.041          & 0.024 $\pm$ 0.003           \\
      SPIDER     & 0.444 $\pm$ 0.341          & 0.201 $\pm$ 0.113          & 1.104 $\pm$ 0.599          & 0.157 $\pm$ 0.019             & 0.191 $\pm$ 0.225          & 0.036 $\pm$ 0.011           \\
      \bottomrule
    \end{tabular}}
\end{table}

\begin{table}[t]
  \caption{Ours: tracking metrics on the benchmark set.}
  \label{tab:main_results_ours}
  \centering
  \scriptsize
  \resizebox{\linewidth}{!}{
  \begin{tabular}{@{}cccccccc@{}}
    \toprule
    Dataset & Clip & Progress $\uparrow$ & Obj pos (m) $\downarrow$ & Obj rot (rad) $\downarrow$ & Finger err (rad) $\downarrow$ & Contact F1 $\uparrow$ & Contact pt (m) $\downarrow$ \\
    \midrule
    \multirow{5}{*}{GRAB} & \texttt{cubemedium\_offhand} & 0.994 & 0.023 & 0.226 & 0.166 & 0.765 & 0.015 \\
     & \texttt{cylindermedium\_offhand} & 1.000 & 0.043 & 0.349 & 0.186 & 0.693 & 0.016 \\
     & \texttt{hammer\_use} & 0.984 & 0.026 & 0.175 & 0.195 & 0.875 & 0.014 \\
     & \texttt{waterbottle\_offhand} & 1.000 & 0.018 & 0.141 & 0.159 & 0.745 & 0.012 \\
     & \texttt{wineglass\_offhand} & 1.000 & 0.029 & 0.200 & 0.170 & 0.915 & 0.011 \\
    \midrule
    \multirow{4}{*}{ARCTIC} & \texttt{box\_use} & 1.000 & 0.020 & 0.110 & 0.152 & 0.865 & 0.021 \\
     & \texttt{mixer\_use} & 0.853 & 0.026 & 0.231 & 0.167 & 0.653 & 0.023 \\
     & \texttt{notebook\_use} & 0.993 & 0.030 & 0.234 & 0.160 & 0.731 & 0.023 \\
     & \texttt{waffleiron\_use} & 0.931 & 0.031 & 0.262 & 0.148 & 0.766 & 0.020 \\
    \midrule
    \multirow{4}{*}{DexterHand} & \texttt{Cuboid\_00} & 0.745 & 0.029 & 0.381 & 0.152 & 0.766 & 0.020 \\
     & \texttt{Cuboid\_01} & 0.987 & 0.025 & 0.400 & 0.147 & 0.781 & 0.014 \\
     & \texttt{Cylinder} & 0.928 & 0.018 & 0.349 & 0.150 & 0.856 & 0.021 \\
     & \texttt{Ring} & 0.272 & 0.024 & 0.474 & 0.169 & 0.789 & 0.025 \\
    \midrule
    GRAB & Avg & 0.996 & 0.028 & 0.218 & 0.175 & 0.798 & 0.013 \\
    ARCTIC & Avg & 0.944 & 0.027 & 0.209 & 0.157 & 0.754 & 0.022 \\
    DexterHand & Avg & 0.733 & 0.024 & 0.401 & 0.155 & 0.798 & 0.020 \\
    Overall & Avg & 0.899 & 0.026 & 0.272 & 0.163 & 0.784 & 0.018 \\
    \bottomrule
  \end{tabular}}
\end{table}

Table~\ref{tab:baselines} reports our main experiment: a baseline comparison under the fixed 5000-update budget. Table~\ref{tab:main_results_ours} shows ConTrack results per clip across tiers. ConTrack reaches near-complete progress on GRAB and strong progress on ARCTIC, while DexterHand remains challenging within the same budget. ManipTrans attains the lowest object pose error but drifts in finger motion and contacts. DexMachina preserves finger motion and contact fidelity more closely but makes limited progress under the same budget. SPIDER stays close in kinematics but struggles to maintain object tracking once contact dynamics dominate. ConTrack narrows this gap by adapting the task--style balance online and by stabilizing long-horizon learning through mid-trajectory resets.

\paragraph{Failure analysis for DexterHand Ring.}
DexterHand Ring is the hardest clip under our fixed budget and failures concentrate in a rotation-dominant phase rather than emerging from gradual drift. With longer training the clip becomes feasible under the same thresholds. See appendix for extended-budget results and termination statistics.

\subsection{Ablations}
\subsubsection{Adaptive vs.\ Fixed Task--Style Mixing}
\begin{figure}[t]
  \centering
  \includegraphics[width=0.8\linewidth]{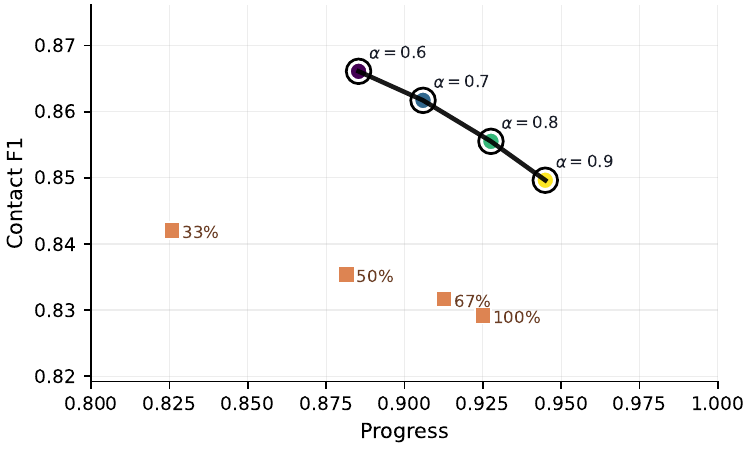}
  \caption{Task--style trade-off controlled by the target ratio $\alpha$. Each round point is a policy trained with a fixed $\alpha$, and each square point is a policy trained with a fixed task--style mixing weight, with task weight annotated. Sweeping $\alpha$ traces an empirical Pareto-optimal frontier between progress and contact fidelity, and fixed reward mixing baselines fall inside this frontier.}
  \label{fig:tradeoff_alpha_sweep}
\end{figure}

\begin{table}[t]
  \caption{Ablation for adaptive vs.\ fixed task--style mixing.}
  \label{tab:ablation_cmdp}
  \centering
  \scriptsize
  \resizebox{\linewidth}{!}{
  \begin{tabular}{@{}ccccccc@{}}
    \toprule
    Variant & Progress $\uparrow$ & Obj pos (m) $\downarrow$ & Obj rot (rad) $\downarrow$ & Finger err (rad) $\downarrow$ & Contact F1 $\uparrow$ & Contact pt (m) $\downarrow$ \\
    \midrule
    Adaptive (ours) & \textbf{0.899} $\pm$ 0.195 & \textbf{0.026} $\pm$ 0.006 & 0.272 $\pm$ 0.105 & 0.163 $\pm$ 0.014 & \textbf{0.784} $\pm$ 0.072 & \textbf{0.018} $\pm$ 0.005 \\
    Fixed task & 0.764 $\pm$ 0.172 & 0.026 $\pm$ 0.004 & \textbf{0.250} $\pm$ 0.031 & \textbf{0.157} $\pm$ 0.021 & 0.679 $\pm$ 0.131 & 0.022 $\pm$ 0.004 \\
    Fixed 1:1 & 0.868 $\pm$ 0.107 & 0.029 $\pm$ 0.008 & 0.297 $\pm$ 0.029 & 0.165 $\pm$ 0.028 & 0.701 $\pm$ 0.106 & 0.023 $\pm$ 0.002 \\
    \bottomrule
  \end{tabular}}
\end{table}

Table~\ref{tab:ablation_cmdp} shows that adaptive mixing improves progress while preserving contact fidelity compared to fixed mixing variants. Fig.~\ref{fig:tradeoff_alpha_sweep} instantiates this trade-off and plots each trained policy as a point in the plane of progress and contact F1. For a policy $\pi$, let $g(\pi)\in[0,1]$ denote its mean progress and let $f(\pi)\in[0,1]$ denote its mean contact F1 against the reference contact annotations. We say $\pi_1$ dominates $\pi_2$ if $g(\pi_1)\ge g(\pi_2)$ and $f(\pi_1)\ge f(\pi_2)$ with at least one strict inequality. Sweeping the target ratio $\alpha$ traces an empirical Pareto frontier across the tested range. In our sweep, fixed reward mixing policies are dominated by points on this frontier, which means that for some achieved progress level, the adaptive update attains higher naturalness as measured by contact fidelity under the same budget.

\subsubsection{Reset Library Ablation}
\begin{figure}[t]
  \centering
  \includegraphics[width=\linewidth]{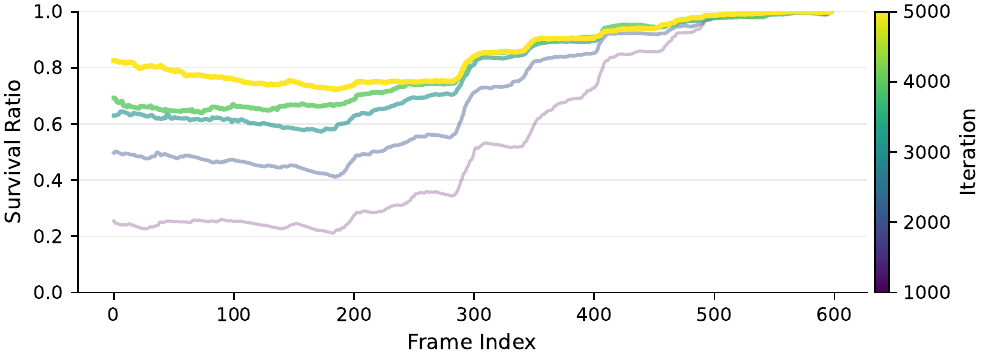}
  \caption{Reset library dynamics over the 5000 training updates. As the policy learns to track further into the clip, the set of reachable mid-trajectory states expands toward earlier frames, allowing resets to move backward and concentrate learning on the remaining difficult segments.}
  \label{fig:reset_histogram}
\end{figure}

\begin{table}[t]
  \caption{Ablation for reset library trick.}
  \label{tab:ablation_reset}
  \centering
  \scriptsize
  \resizebox{\linewidth}{!}{
  \begin{tabular}{@{}ccccccc@{}}
    \toprule
    Variant & Progress $\uparrow$ & Obj pos (m) $\downarrow$ & Obj rot (rad) $\downarrow$ & Finger err (rad) $\downarrow$ & Contact F1 $\uparrow$ & Contact pt (m) $\downarrow$ \\
    \midrule
    Adaptive library (Ours) & \textbf{0.899} $\pm$ 0.195 & \textbf{0.026} $\pm$ 0.006 & \textbf{0.272} $\pm$ 0.105 & 0.163 $\pm$ 0.014 & \textbf{0.784} $\pm$ 0.072 & 0.018 $\pm$ 0.005 \\
    Reset from start & 0.700 $\pm$ 0.270 & 0.039 $\pm$ 0.010 & 0.370 $\pm$ 0.045 & 0.168 $\pm$ 0.011 & 0.739 $\pm$ 0.088 & \textbf{0.017} $\pm$ 0.002 \\
    Reset uniform & 0.727 $\pm$ 0.348 & 0.033 $\pm$ 0.010 & 0.298 $\pm$ 0.080 & \textbf{0.152} $\pm$ 0.016 & 0.714 $\pm$ 0.119 & 0.021 $\pm$ 0.004 \\
    \bottomrule
  \end{tabular}}
\end{table}

Table~\ref{tab:ablation_reset} shows that the reset library improves progress under the same budget compared to always resetting from the first frame and to a uniform mid-clip reset. Fig.~\ref{fig:reset_histogram} tracks the distribution of sampled reset frames throughout training. Early updates concentrate resets near the end of the clip where short-horizon tracking is feasible. As learning expands the reachable suffix, resets move earlier, so training continually targets the current transition point between success and failure.

\subsubsection{Contact Prior Reward Ablation}

\begin{table}[t]
  \caption{Ablation for contact prior reward.}
  \label{tab:ablation_contact}
  \centering
  \scriptsize
  \resizebox{\linewidth}{!}{
    \begin{tabular}{@{}ccccccc@{}}
      \toprule
      Variant                     & Progress $\uparrow$        & Obj pos (m) $\downarrow$   & Obj rot (rad) $\downarrow$ & Finger err (rad) $\downarrow$ & Contact F1 $\uparrow$      & Contact pt (m) $\downarrow$ \\
      \midrule
      Full (Ours)                 & \textbf{0.899} $\pm$ 0.195 & \textbf{0.026} $\pm$ 0.006 & \textbf{0.272} $\pm$ 0.105 & 0.163 $\pm$ 0.014             & \textbf{0.784} $\pm$ 0.072 & \textbf{0.018} $\pm$ 0.005  \\
      w/o contact reward          & 0.861 $\pm$ 0.173          & 0.032 $\pm$ 0.010          & 0.320 $\pm$ 0.044          & 0.168 $\pm$ 0.019             & 0.699 $\pm$ 0.092          & 0.020 $\pm$ 0.005           \\
      w/o contact distance reward & 0.868 $\pm$ 0.057          & 0.032 $\pm$ 0.011          & 0.288 $\pm$ 0.071          & \textbf{0.149} $\pm$ 0.016    & 0.753 $\pm$ 0.089          & 0.023 $\pm$ 0.005           \\
      \bottomrule
    \end{tabular}}
\end{table}

Table~\ref{tab:ablation_contact} indicates that contact priors provide complementary supervision for contact-rich tracking. Removing either contact term reduces progress or contact accuracy, suggesting that the priors help resolve contact dynamics that are not fully specified by object pose errors alone.

\subsection{Real-world Feasibility}
\label{sec:hardware}

\begin{figure}[t]
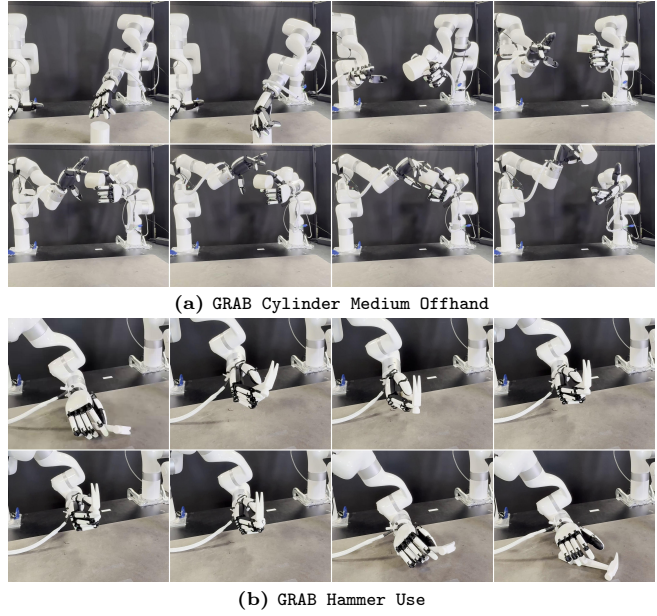

  \centering
  \begin{subfigure}{\linewidth}
    \centering
    \includegraphics[width=0.7\linewidth]{figures/real_grab_s1_cylindermedium_offhand_1.pdf}
    \caption{\texttt{GRAB Cylinder Medium Offhand}}
  \end{subfigure}
  \begin{subfigure}{\linewidth}
    \centering
    \includegraphics[width=0.7\linewidth]{figures/real_grab_s1_hammer_use_1.pdf}
    \caption{\texttt{GRAB Hammer Use}}
  \end{subfigure}

  \caption{Real-world experiment snapshots on the bimanual xArm7+xHand platform.}
  \label{fig:clip_montages_real_main}
\end{figure}

We study real-world feasibility on a tabletop bimanual platform with two xArm7 arms and two xHands. For each clip, we stream the policy-predicted joint references from the simulator to a separate real-time controller over TCP. Each side tracks $7$ arm joints and $12$ hand joints, with outer reference updates at $0.15$~s and an internal arm control loop running at $250$~Hz. Fig.~\ref{fig:clip_montages_real_main} shows representative executions, and additional experiments are shown in appendix.

\section{Conclusion}
\label{sec:conclusion}

\paragraph{Limitations.}
Our constrained formulation uses an online normalization based on a running maximum of task return estimates, which provides stable scaling in practice but does not strictly guarantee satisfaction of the constraint. Contact priors rely on the availability and quality of contact annotations, and tracking remains challenging for the hardest in-hand manipulation clips under a limited interaction budget. Finally, our hardware study currently focuses on executing joint commands on the real arm--hand platform. Extending the system with richer perception and tighter sim-to-real alignment could further broaden robustness in real-world deployment.

In summary, we presented ConTrack, a constrained reference-tracking reinforcement learning framework for long-horizon, contact-rich hand--object interaction in physics simulation. ConTrack separates task success from style fidelity and controls their trade-off online through a scalar dual controller. To address the instability of learning long clips from the first frame, ConTrack maintains an adaptive mid-trajectory reset library that refreshes entries from policy-reachable simulator states, and samples start frames to focus learning on difficult segments. Across bimanual interaction and in-hand manipulation benchmarks under a fixed training budget, ConTrack improves progress and contact fidelity while retaining accurate object tracking, and exposes an interpretable target ratio that traces a smooth task--style trade-off frontier.

\clearpage
\subsubsection*{Acknowledgment}
This work was supported, in part, by NSF CAREER Award IIS-2240014 and NSF CCF-2112665 (TILOS).

\bibliographystyle{splncs04}
\bibliography{main}

\begin{thebibliography}{10}
\providecommand{\url}[1]{\texttt{#1}}
\providecommand{\urlprefix}{URL }
\providecommand{\doi}[1]{https://doi.org/#1}

\bibitem{achiam2017-cpo}
Achiam, J., Held, D., Tamar, A., Abbeel, P.: Constrained policy optimization.
  In: Precup, D., Teh, Y.W. (eds.) Proceedings of the 34th International
  Conference on Machine Learning. Proceedings of Machine Learning Research,
  vol.~70, pp. 22--31. PMLR (06--11 Aug 2017)

\bibitem{barreiros2025-lbm}
Barreiros, J., Beaulieu, A., Bhat, A., Cory, R., Cousineau, E., Dai, H., Fang,
  C.H., Hashimoto, K., Irshad, M.Z., Itkina, M., et~al.: A careful examination
  of large behavior models for multitask dexterous manipulation. arXiv preprint
  arXiv:2507.05331  (2025)

\bibitem{brahmbhatt2020contactpose}
Brahmbhatt, S., Tang, C., Twigg, C.D., Kemp, C.C., Hays, J.: {ContactPose}: A
  dataset of grasps with object contact and hand pose. In: The European
  Conference on Computer Vision (ECCV) (August 2020)

\bibitem{cai2025-in-n-on}
Cai, X., Qiu, R.Z., Chen, G., Wei, L., Liu, I., Huang, T., Cheng, X., Wang, X.:
  In-n-on: Scaling egocentric manipulation with in-the-wild and on-task data.
  arXiv preprint arXiv:2511.15704  (2025)

\bibitem{carpentier2019-pinocchio}
Carpentier, J., Saurel, G., Buondonno, G., Mirabel, J., Lamiraux, F., Stasse,
  O., Mansard, N.: The pinocchio c++ library -- a fast and flexible
  implementation of rigid body dynamics algorithms and their analytical
  derivatives. In: IEEE International Symposium on System Integrations (SII)
  (2019)

\bibitem{chao2021dex}
Chao, Y.W., Yang, W., Xiang, Y., Molchanov, P., Handa, A., Tremblay, J.,
  Narang, Y.S., {Van Wyk}, K., Iqbal, U., Birchfield, S., Kautz, J., Fox, D.:
  {DexYCB}: A benchmark for capturing hand grasping of objects. In: IEEE/CVF
  Conference on Computer Vision and Pattern Recognition (CVPR) (2021)

\bibitem{chen2024-object}
Chen, Y., Wang, C., Yang, Y., Liu, C.K.: Object-centric dexterous manipulation
  from human motion data. In: CoRL (2024)

\bibitem{cheng2024-opentv}
Cheng, X., Li, J., Yang, S., Yang, G., Wang, X.: Open-television: Teleoperation
  with immersive active visual feedback. arXiv preprint arXiv:2407.01512
  (2024)

\bibitem{chi2024-umi}
Chi, C., Xu, Z., Pan, C., Cousineau, E., Burchfiel, B., Feng, S., Tedrake, R.,
  Song, S.: Universal manipulation interface: In-the-wild robot teaching
  without in-the-wild robots. arXiv preprint arXiv:2402.10329  (2024)

\bibitem{ecoffet2019-goexplore}
Ecoffet, A., Huizinga, J., Lehman, J., Stanley, K.O., Clune, J.: Go-explore: a
  new approach for hard-exploration problems. arXiv preprint arXiv:1901.10995
  (2019)

\bibitem{fan2023arctic}
Fan, Z., Taheri, O., Tzionas, D., Kocabas, M., Kaufmann, M., Black, M.J.,
  Hilliges, O.: {ARCTIC}: A dataset for dexterous bimanual hand-object
  manipulation. In: Proceedings IEEE Conference on Computer Vision and Pattern
  Recognition (CVPR) (2023)

\bibitem{fu2025gigahands}
Fu, R., Zhang, D., Jiang, A., Fu, W., Funk, A., Ritchie, D., Sridhar, S.:
  Gigahands: A massive annotated dataset of bimanual hand activities. In:
  Proceedings of the Computer Vision and Pattern Recognition Conference. pp.
  17461--17474 (2025)

\bibitem{geng2025roboverse}
Geng, H., Wang, F., Wei, S., Li, Y., Wang, B., An, B., Lou, H., Cheng, C.T.,
  Li, P., Chen, H., Liang, Y., Qian, Y., Mao, J., Wan, W., Geng, Y., Zhang, M.,
  Lyu, J., Zhao, S., Zhang, J., Xu, C., Zhang, J., Zhao, C., Lu, H., Ding, Y.,
  Gong, R., Wang, Y., Kuang, Y., Wu, R., Jia, B., Dong, H., Huang, S., Wang,
  Y., Malik, J., Abbeel, P.: {RoboVerse: A Unified Platform, Benchmark and
  Dataset for Scalable and Generalizable Robot Learning}. In: Proceedings of
  Robotics: Science and Systems. Los Angeles, CA, USA (June 2025).
  \doi{10.15607/RSS.2025.XXI.022},
  \url{https://roboticsproceedings.org/rss21/p022.html}, arXiv:2504.18904

\bibitem{handa2020-dexpilot}
Handa, A., Van~Wyk, K., Yang, W., Liang, J., Chao, Y.W., Wan, Q., Birchfield,
  S., Ratliff, N., Fox, D.: Dexpilot: Vision-based teleoperation of dexterous
  robotic hand-arm system. In: ICRA (2020)

\bibitem{hsieh2025dexman}
Hsieh, J., Tu, K.H., Hung, K.H., Ke, T.W.: Dexman: Learning bimanual dexterous
  manipulation from human and generated videos. arXiv preprint arXiv:2510.08475
   (2025)

\bibitem{intelligence2025pi_05}
Intelligence, P., Black, K., Brown, N., Darpinian, J., Dhabalia, K., Driess,
  D., Esmail, A., Equi, M., Finn, C., Fusai, N., et~al.: pi0.5: a
  vision-language-action model with open-world generalization. arXiv preprint
  arXiv:2504.16054  (2025)

\bibitem{ishihara2025-constraints}
Ishihara, Y., Takasugi, N., Kawakami, K., Kinoshita, M., Aoyama, K.:
  Constraints as rewards: Reinforcement learning for robots without reward
  functions. arXiv preprint arXiv:2501.04228  (2025)

\bibitem{jian2023affordpose}
Jian, J., Liu, X., Li, M., Hu, R., Liu, J.: Affordpose: A large-scale dataset
  of hand-object interactions with affordance-driven hand pose. In: Proceedings
  of the IEEE/CVF International Conference on Computer Vision (ICCV). pp.
  14713--14724 (October 2023)

\bibitem{jiang2025gsworld}
Jiang, G., Chang, H., Qiu, R.Z., Liang, Y., Ji, M., Zhu, J., Dong, Z., Zou, X.,
  Wang, X.: {GSWorld: Closed-Loop Photorealistic Simulation Suite for Robot
  Manipulation}. In: 2026 IEEE International Conference on Robotics and
  Automation (ICRA). pp. 384--391. IEEE (June 2026).
  \doi{10.1109/ICRA57385.2026.11697084},
  \url{https://doi.org/10.1109/ICRA57385.2026.11697084}, arXiv:2510.20813

\bibitem{jiang2026crosshand}
Jiang, G., Liang, Y., Ye, J., Huang, J.Y., Jing, C., Duan, R., Abbeel, P.,
  Wang, X., Zou, X.: {Cross-Hand Latent Representation for
  Vision-Language-Action Models}. In: Proceedings of the IEEE/CVF Conference on
  Computer Vision and Pattern Recognition (CVPR). pp. 13496--13507 (June 2026),
  \url{https://openaccess.thecvf.com/content/CVPR2026/html/Jiang_Cross-Hand_Latent_Representation_for_Vision-Language-Action_Models_CVPR_2026_paper.html},
  arXiv:2603.10158

\bibitem{jiang2023chairs}
Jiang, N., Liu, T., Cao, Z., Cui, J., Chen, Y., Wang, H., Zhu, Y., Huang, S.:
  Full-body articulated human-object interaction. In: ICCV (2023)

\bibitem{kareer2025-egomimic}
Kareer, S., Patel, D., Punamiya, R., Mathur, P., Cheng, S., Wang, C., Hoffman,
  J., Xu, D.: Egomimic: Scaling imitation learning via egocentric video. In:
  ICRA. IEEE (2025)

\bibitem{kareer2025-emergence}
Kareer, S., Pertsch, K., Darpinian, J., Hoffman, J., Xu, D., Levine, S., Finn,
  C., Nair, S.: Emergence of human to robot transfer in vision-language-action
  models. arXiv preprint arXiv:2512.22414  (2025)

\bibitem{kim2024parahome}
Kim, J., Kim, J., Na, J., Joo, H.: { ParaHome: Parameterizing Everyday Home
  Activities Towards 3D Generative Modeling of Human-Object Interactions }. In:
  2025 IEEE/CVF Conference on Computer Vision and Pattern Recognition (CVPR).
  pp. 1816--1828. IEEE Computer Society (2025)

\bibitem{li2025-maniptrans}
Li, K., Li, P., Liu, T., Li, Y., Huang, S.: Maniptrans: Efficient dexterous
  bimanual manipulation transfer via residual learning. In: CVPR (2025)

\bibitem{liang2026fingr}
Liang, Y., Peng, Q., Kim, M., Wang, X.: {FINGR: Learning Dexterous Hand Control
  for Real-World Rubik's Cube Solving}. arXiv preprint arXiv:2609.33973 (2026),
  \url{https://arxiv.org/abs/2609.33973}

\bibitem{liang2026dextercap}
Liang, Y., Xu, S., Zhang, Y., Zhan, B., Zhang, H., Liu, L.: {DexterCap:
  Affordable and Automated Capture of Complex Hand-Object Interactions}.
  Computer Graphics Forum  \textbf{45}(2),  e70330 (2026).
  \doi{10.1111/cgf.70330}, \url{https://doi.org/10.1111/cgf.70330},
  arXiv:2601.05844

\bibitem{liao2025beyondmimic}
Liao, Q., Truong, T.E., Huang, X., Gao, Y., Tevet, G., Sreenath, K., Liu, C.K.:
  Beyondmimic: From motion tracking to versatile humanoid control via guided
  diffusion. arXiv preprint arXiv:2508.08241  (2025)

\bibitem{liu2024rdt}
Liu, S., Wu, L., Li, B., Tan, H., Chen, H., Wang, Z., Xu, K., Su, H., Zhu, J.:
  Rdt-1b: a diffusion foundation model for bimanual manipulation. arXiv
  preprint arXiv:2410.07864  (2024)

\bibitem{liu2025-dextrack}
Liu, X., Adalibieke, J., Han, Q., Qin, Y., Yi, L.: Dextrack: Towards
  generalizable neural tracking control for dexterous manipulation from human
  references. In: ICLR (2025)

\bibitem{lu2025humoto}
Lu, J., Huang, C.H.P., Bhattacharya, U., Huang, Q., Zhou, Y.: Humoto: A 4d
  dataset of mocap human object interactions. In: Proceedings of the IEEE/CVF
  International Conference on Computer Vision (ICCV). pp. 10886--10897 (October
  2025)

\bibitem{luo2026-being}
Luo, H., Wang, Y., Zhang, W., Zheng, S., Xi, Z., Xu, C., Xu, H., Yuan, H.,
  Zhang, C., Wang, Y., et~al.: Being-h0. 5: Scaling human-centric robot
  learning for cross-embodiment generalization. arXiv preprint arXiv:2601.12993
   (2026)

\bibitem{luo2025sonic}
Luo, Z., Yuan, Y., Wang, T., Li, C., Chen, S., Castaneda, F., Cao, Z.A., Li,
  J., Minor, D., Ben, Q., et~al.: Sonic: Supersizing motion tracking for
  natural humanoid whole-body control. arXiv preprint arXiv:2511.07820  (2025)

\bibitem{mandi2025-dexmachina}
Mandi, Z., Hou, Y., Fox, D., Narang, Y., Mandlekar, A., Song, S.: Dexmachina:
  Functional retargeting for bimanual dexterous manipulation. arXiv preprint
  arXiv:2505.24853  (2025)

\bibitem{pan2025-spider}
Pan, C., Wang, C., Qi, H., Liu, Z., Bharadhwaj, H., Sharma, A., Wu, T., Shi,
  G., Malik, J., Hogan, F.: Spider: Scalable physics-informed dexterous
  retargeting. arXiv preprint arXiv:2511.09484  (2025)

\bibitem{peng2026fact}
Peng, Q., Liang, Y., Yan, R., Hansen, N., Wang, X.: {FACT: Failure-Aware Causal
  Training for World-Action Models}. In: Conference on Robot Learning (CoRL)
  (2026), \url{https://arxiv.org/abs/2608.10232}, arXiv:2608.10232

\bibitem{peng2018deepmimic}
Peng, X.B., Abbeel, P., Levine, S., van~de Panne, M.: Deepmimic: Example-guided
  deep reinforcement learning of physics-based character skills. ACM Trans.
  Graph.  \textbf{37}(4),  143:1--143:14 (Jul 2018)

\bibitem{peng2022ase}
Peng, X.B., Guo, Y., Halper, L., Levine, S., Fidler, S.: Ase: Large-scale
  reusable adversarial skill embeddings for physically simulated characters.
  ACM Trans. Graph.  \textbf{41}(4) (Jul 2022)

\bibitem{qin2023-anyteleop}
Qin, Y., Yang, W., Huang, B., Van~Wyk, K., Su, H., Wang, X., Chao, Y.W., Fox,
  D.: Anyteleop: A general vision-based dexterous robot arm-hand teleoperation
  system. arXiv preprint arXiv:2307.04577  (2023)

\bibitem{qiu2025-humanoid}
Qiu, R.Z., Yang, S., Cheng, X., Chawla, C., Li, J., He, T., Yan, G., Yoon,
  D.J., Hoque, R., Paulsen, L., et~al.: Humanoid policy\~{} human policy. In:
  CoRL (2025)

\bibitem{resnick2018-backplay}
Resnick, C., Raileanu, R., Kapoor, S., Peysakhovich, A., Cho, K., Bruna, J.:
  Backplay:" man muss immer umkehren". arXiv preprint arXiv:1807.06919  (2018)

\bibitem{schulman2017-ppo}
Schulman, J., Wolski, F., Dhariwal, P., Radford, A., Klimov, O.: Proximal
  policy optimization algorithms. arXiv preprint arXiv:1707.06347  (2017)

\bibitem{song2025-opentouch}
Song, Y.R., Li, J., Fu, R., Murphy, D., Zhou, K., Shiv, R., Li, Y., Xiong, H.,
  Owens, C.E., Du, Y., et~al.: Opentouch: Bringing full-hand touch to
  real-world interaction. arXiv preprint arXiv:2512.16842  (2025)

\bibitem{taheri2020grab}
Taheri, O., Ghorbani, N., Black, M.J., Tzionas, D.: {GRAB}: A dataset of
  whole-body human grasping of objects. In: European Conference on Computer
  Vision (ECCV) (2020)

\bibitem{spiritai2026-spiritv15}
Team, S.A.: Spirit-v1.5: Clean data is the enemy of great robot foundation
  models. Spirit AI Blog  (2026)

\bibitem{tessler2024maskedmimic}
Tessler, C., Guo, Y., Nabati, O., Chechik, G., Peng, X.B.: Maskedmimic: Unified
  physics-based character control through masked motion inpainting. ACM
  Transactions on Graphics (TOG)  (2024)

\bibitem{tessler2018-rcpo}
Tessler, C., Mankowitz, D.J., Mannor, S.: Reward constrained policy
  optimization. arXiv preprint arXiv:1805.11074  (2018)

\bibitem{wen2025-constrained}
Wen, K., Li, C., He, J., Hutter, M.: Constrained style learning from imperfect
  demonstrations under task optimality. arXiv preprint arXiv:2507.09371  (2025)

\bibitem{wen2025-grdexter}
Wen, R., Chen, G., Cui, Z., Du, M., Gou, Y., Han, Z., Huang, L., Lei, M., Li,
  Y., Li, Z., et~al.: Gr-dexter technical report. arXiv preprint
  arXiv:2512.24210  (2025)

\bibitem{xu2025dexumi}
Xu, M., Zhang, H., Hou, Y., Xu, Z., Fan, L., Veloso, M., Song, S.: Dexumi:
  Using human hand as the universal manipulation interface for dexterous
  manipulation. arXiv preprint arXiv:2505.21864  (2025)

\bibitem{xu2025dexplore}
Xu, S., Chao, Y.W., Bian, L., Mousavian, A., Wang, Y.X., Gui, L.Y., Yang, W.:
  Dexplore: Scalable neural control for dexterous manipulation from
  reference-scoped exploration. arXiv preprint arXiv:2509.09671  (2025)

\bibitem{yin2025-geometric-retargeting}
Yin, Z.H., Wang, C., Pineda, L., Bodduluri, K., Wu, T., Abbeel, P., Mukadam,
  M.: Geometric retargeting: A principled, ultrafast neural hand retargeting
  algorithm. In: IROS (2025)

\bibitem{ze2025twist}
Ze, Y., Chen, Z., Araujo, J.P., Cao, Z.a., Peng, X.B., Wu, J., Liu, K.: Twist:
  Teleoperated whole-body imitation system. In: Lim, J., Song, S., Park, H.W.
  (eds.) Proceedings of The 9th Conference on Robot Learning. Proceedings of
  Machine Learning Research, vol.~305, pp. 2143--2154. PMLR (27--30 Sep 2025)

\bibitem{zhao2026dexh2r}
Zhao, S., Zhu, X., Chen, Y., Li, C., Xie, L., Zhang, X., Ding, M., Tomizuka,
  M.: Dexh2r: Task-oriented dexterous manipulation from human to robots. arXiv
  preprint arXiv:2411.04428  (2026)

\bibitem{zheng2026egoscale}
Zheng, R., Niu, D., Xie, Y., Wang, J., Xu, M., Jiang, Y., Casta{\~n}eda, F.,
  Hu, F., Tan, Y.L., Fu, L., et~al.: Egoscale: Scaling dexterous manipulation
  with diverse egocentric human data. arXiv preprint arXiv:2602.16710  (2026)

\end{thebibliography}
\clearpage
\appendix
\section{Additional Clip Snapshots}
We additionally benchmark on the Sharpa Wave embodiment. Table~\ref{tab:appendix_sharpa_wave_results} reports the same six metrics as Table~\ref{tab:main_results_ours}.

\begin{figure*}[t]
  \centering
  \begin{subfigure}{\linewidth}
    \centering
    \includegraphics[width=0.49\linewidth]{figures/sim_grab_s1_cubemedium_offhand_1_xhand.pdf}
    \hfill
    \includegraphics[width=0.49\linewidth]{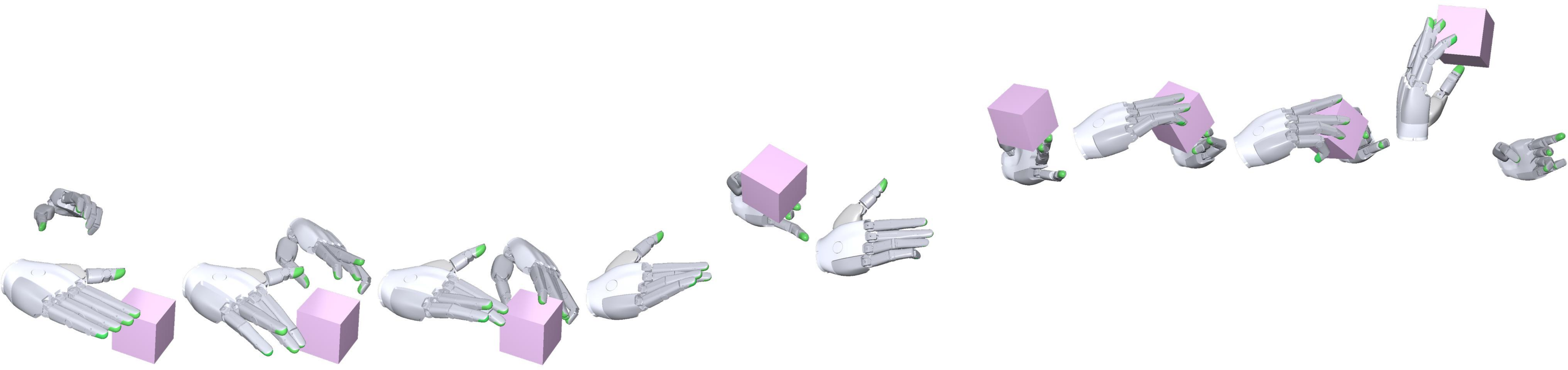}
    \caption{\texttt{GRAB Cube Medium Offhand}}
  \end{subfigure}

  \begin{subfigure}{\linewidth}
    \centering
    \includegraphics[width=0.49\linewidth]{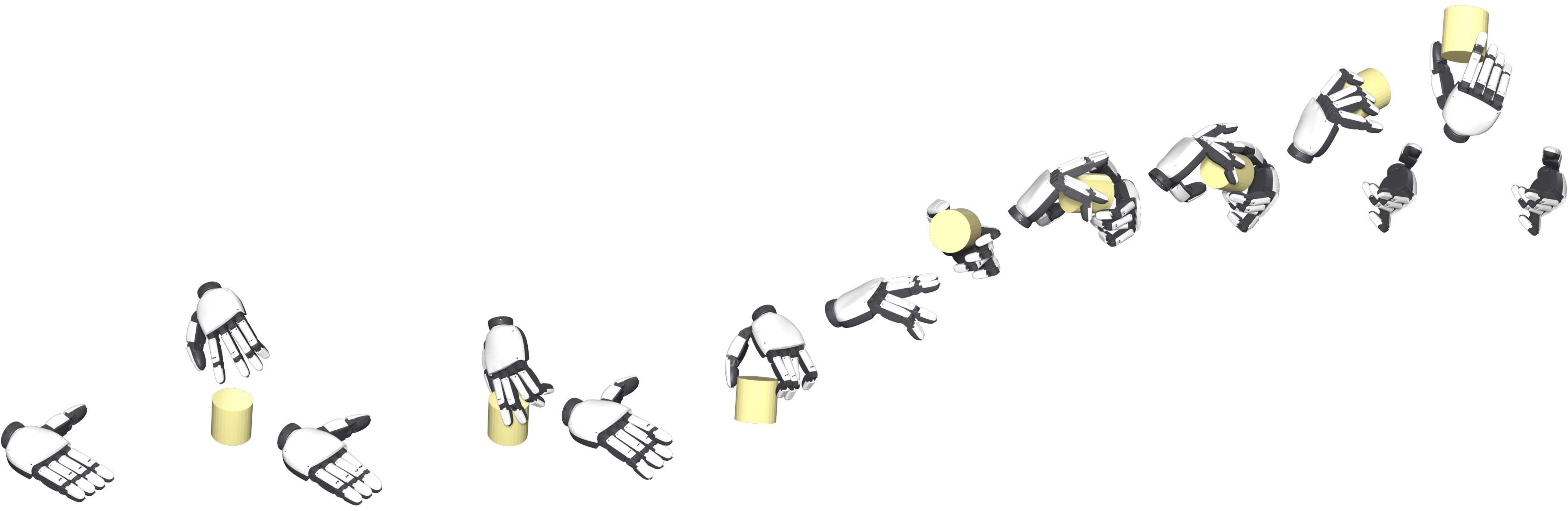}
    \hfill
    \includegraphics[width=0.49\linewidth]{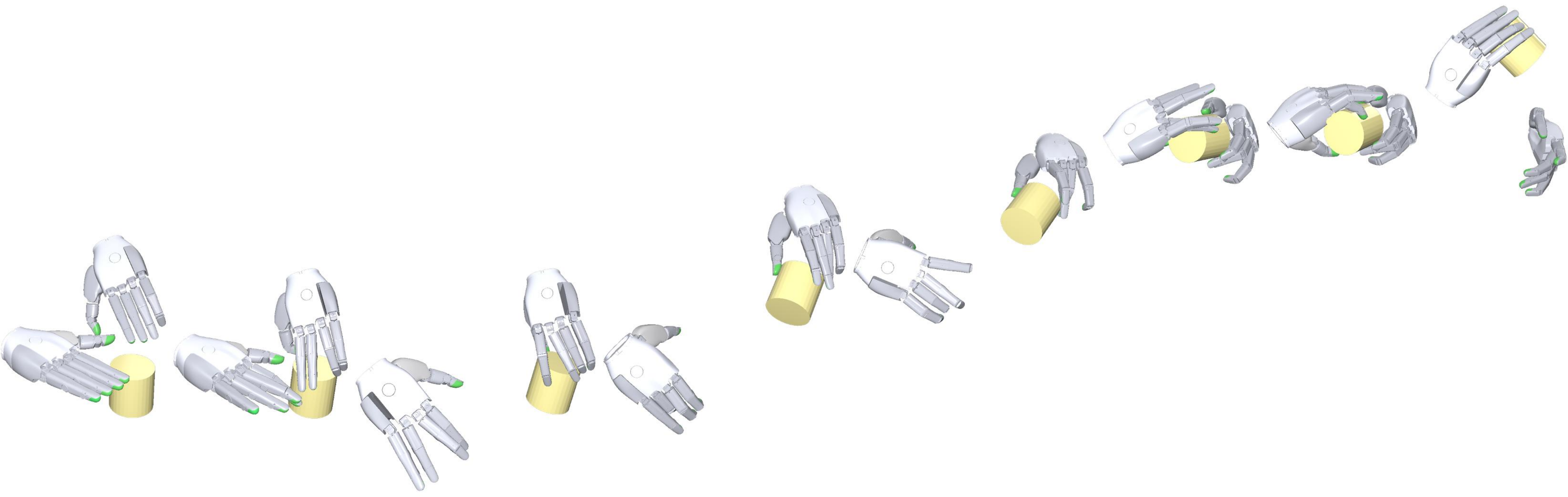}
    \caption{\texttt{GRAB Cylinder Medium Offhand}}
  \end{subfigure}

  \begin{subfigure}{\linewidth}
    \centering
    \includegraphics[width=0.49\linewidth]{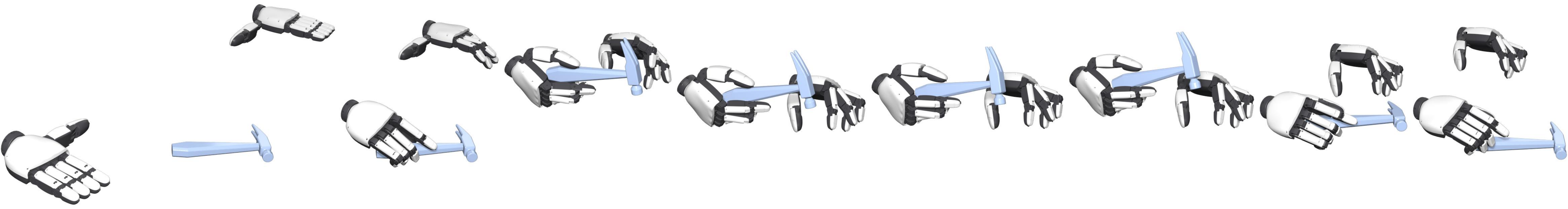}
    \hfill
    \includegraphics[width=0.49\linewidth]{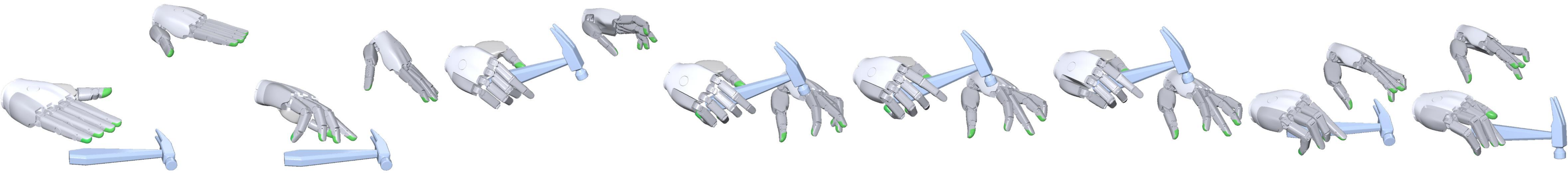}
    \caption{\texttt{GRAB Hammer Use}}
  \end{subfigure}

  \begin{subfigure}{\linewidth}
    \centering
    \includegraphics[width=0.49\linewidth]{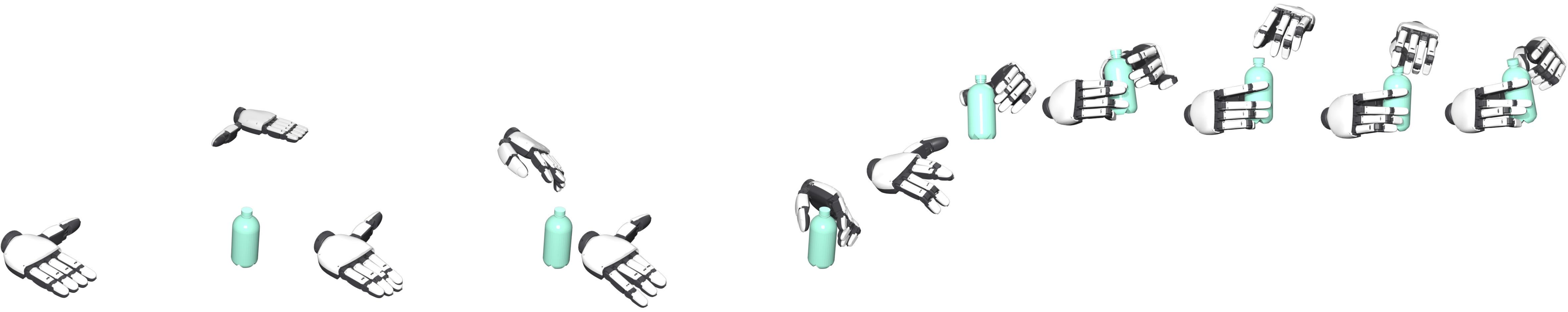}
    \hfill
    \includegraphics[width=0.49\linewidth]{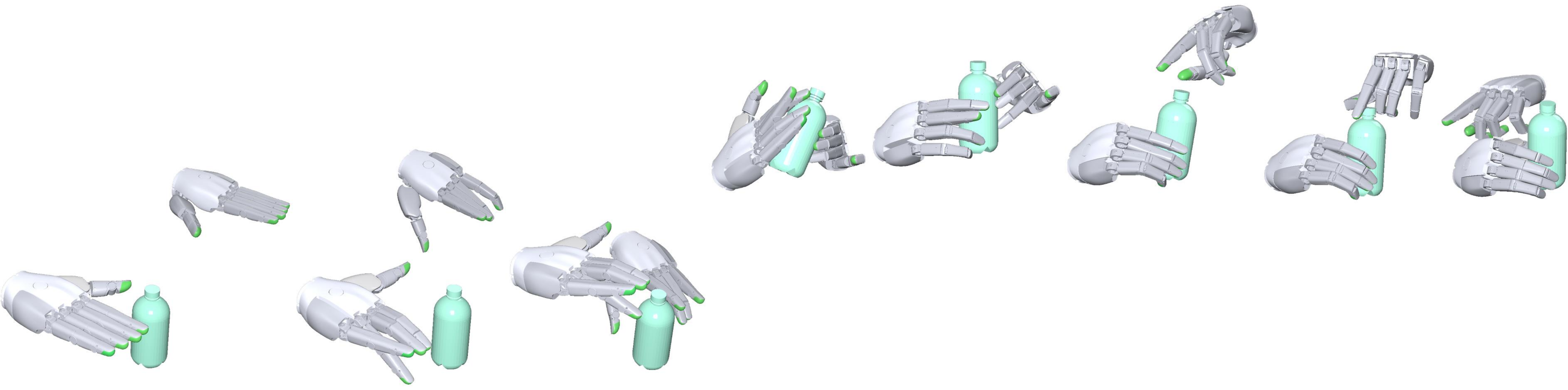}
    \caption{\texttt{GRAB Waterbottle Offhand}}
  \end{subfigure}

  \begin{subfigure}{\linewidth}
    \centering
    \includegraphics[width=0.49\linewidth]{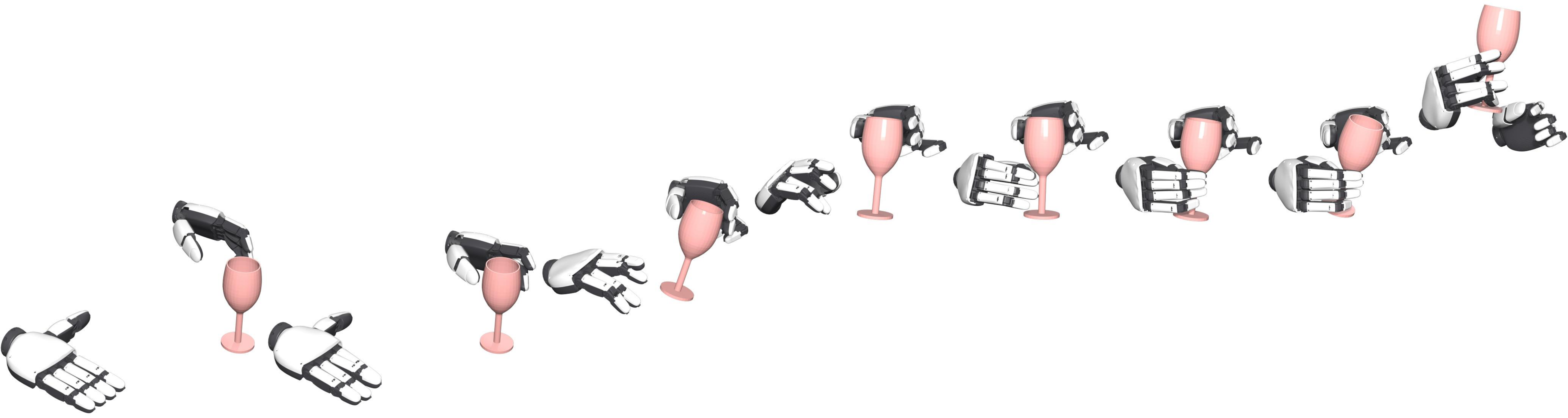}
    \hfill
    \includegraphics[width=0.49\linewidth]{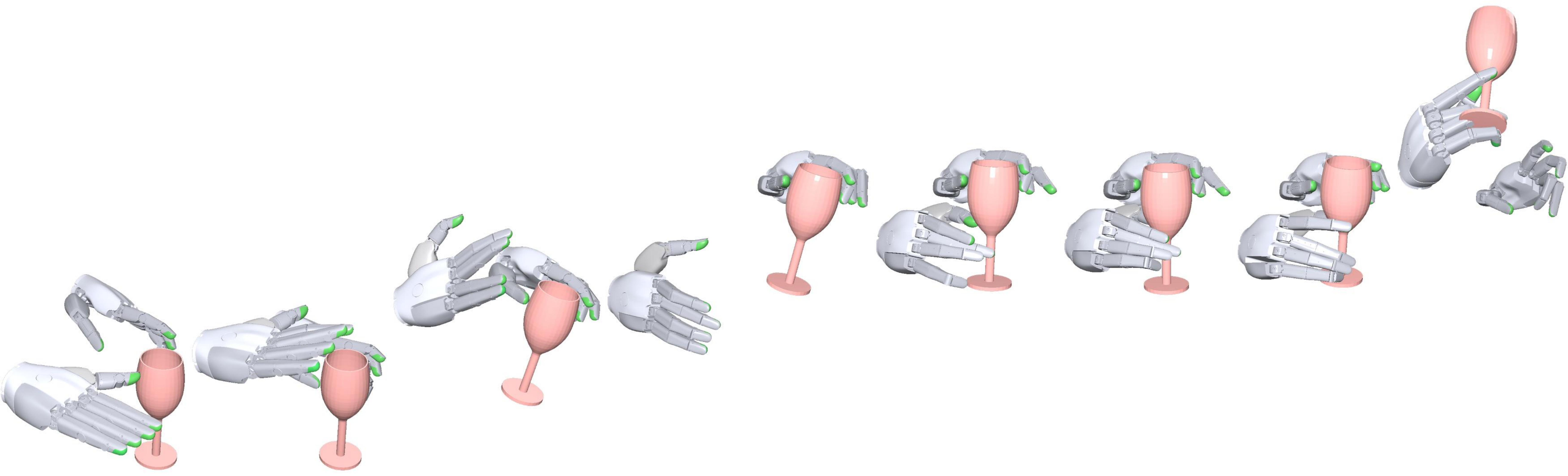}
    \caption{\texttt{GRAB Wineglass Offhand}}
  \end{subfigure}

  \caption{Additional GRAB clips.}
  \label{fig:appendix_sharpa_wave_part1}
\end{figure*}

\begin{figure*}[t]
  \centering
  \begin{subfigure}{\linewidth}
    \centering
    \includegraphics[width=0.49\linewidth]{figures/sim_arctic_s01_box_use_01_xhand.pdf}
    \hfill
    \includegraphics[width=0.49\linewidth]{figures/sim_arctic_s01_box_use_01_sharpa.pdf}
    \caption{\texttt{ARCTIC Box Use}}
  \end{subfigure}

  \begin{subfigure}{\linewidth}
    \centering
    \includegraphics[width=0.49\linewidth]{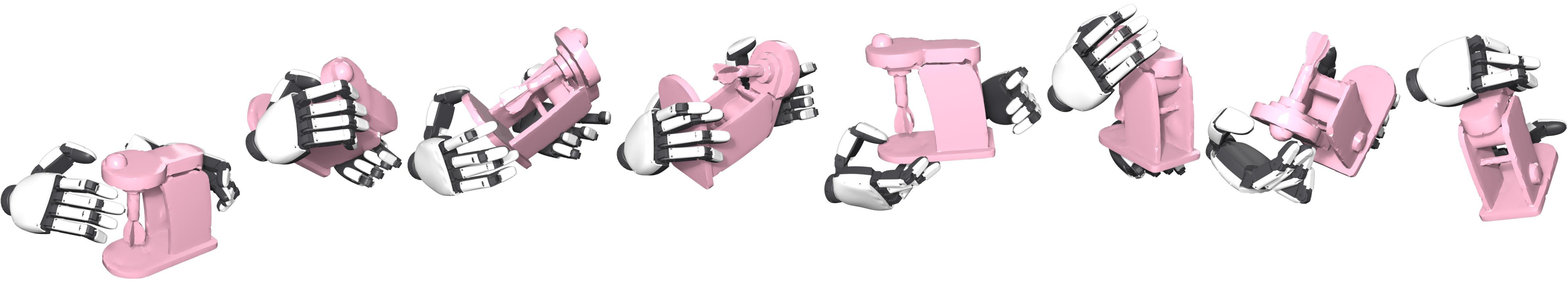}
    \hfill
    \includegraphics[width=0.49\linewidth]{figures/sim_arctic_s01_mixer_use_01_sharpa.pdf}
    \caption{\texttt{ARCTIC Mixer Use}}
  \end{subfigure}

  \begin{subfigure}{\linewidth}
    \centering
    \includegraphics[width=0.49\linewidth]{figures/sim_arctic_s01_notebook_use_01_xhand.pdf}
    \hfill
    \includegraphics[width=0.49\linewidth]{figures/sim_arctic_s01_notebook_use_01_sharpa.pdf}
    \caption{\texttt{ARCTIC Notebook Use}}
  \end{subfigure}

  \begin{subfigure}{\linewidth}
    \centering
    \includegraphics[width=0.49\linewidth]{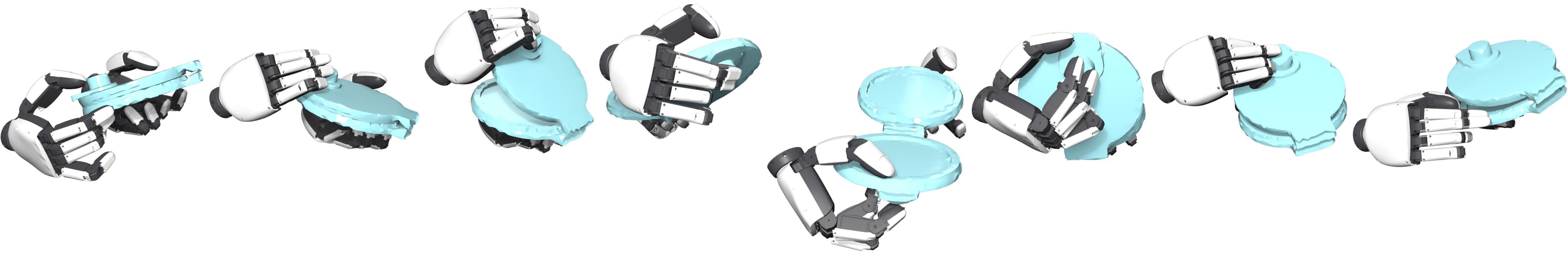}
    \hfill
    \includegraphics[width=0.49\linewidth]{figures/sim_arctic_s01_waffleiron_use_01_sharpa.pdf}
    \caption{\texttt{ARCTIC Waffleiron Use}}
  \end{subfigure}

  \caption{Additional ARCTIC clips.}
  \label{fig:appendix_sharpa_wave_part2}
\end{figure*}

\begin{figure*}[t]
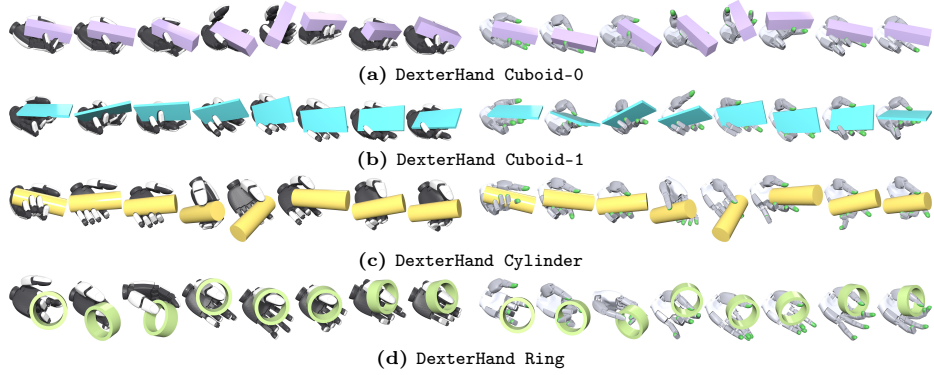

  \centering
  \begin{subfigure}{\linewidth}
    \centering
    \includegraphics[width=0.49\linewidth]{figures/sim_dexterhand_cuboid_00_xhand.pdf}
    \hfill
    \includegraphics[width=0.49\linewidth]{figures/sim_dexterhand_cuboid_00_sharpa.pdf}
    \caption{\texttt{DexterHand Cuboid-0}}
  \end{subfigure}

  \begin{subfigure}{\linewidth}
    \centering
    \includegraphics[width=0.49\linewidth]{figures/sim_dexterhand_cuboid_01_xhand.pdf}
    \hfill
    \includegraphics[width=0.49\linewidth]{figures/sim_dexterhand_cuboid_01_sharpa.pdf}
    \caption{\texttt{DexterHand Cuboid-1}}
  \end{subfigure}

  \begin{subfigure}{\linewidth}
    \centering
    \includegraphics[width=0.49\linewidth]{figures/sim_dexterhand_cylinder_xhand.pdf}
    \hfill
    \includegraphics[width=0.49\linewidth]{figures/sim_dexterhand_cylinder_sharpa.pdf}
    \caption{\texttt{DexterHand Cylinder}}
  \end{subfigure}

  \begin{subfigure}{\linewidth}
    \centering
    \includegraphics[width=0.49\linewidth]{figures/sim_dexterhand_ring_xhand.pdf}
    \hfill
    \includegraphics[width=0.49\linewidth]{figures/sim_dexterhand_ring_sharpa.pdf}
    \caption{\texttt{DexterHand Ring}}
  \end{subfigure}

  \caption{Additional DexterHand clips.}
  \label{fig:appendix_sharpa_wave_part3}
\end{figure*}

More snapshots are shown in Figs.~\ref{fig:appendix_sharpa_wave_part1}, ~\ref{fig:appendix_sharpa_wave_part2}, and~\ref{fig:appendix_sharpa_wave_part3}.

\begin{table}[t]
  \caption{Sharpa Wave: tracking metrics on the benchmark set.}
  \label{tab:appendix_sharpa_wave_results}
  \centering
  \scriptsize
  \resizebox{\linewidth}{!}{
  \begin{tabular}{@{}cccccccc@{}}
    \toprule
    Dataset & Progress $\uparrow$ & Obj pos (m) $\downarrow$ & Obj rot (rad) $\downarrow$ & Finger err (rad) $\downarrow$ & Contact F1 $\uparrow$ & Contact pt (m) $\downarrow$ \\
    \midrule
    GRAB & 0.999 & 0.020 & 0.128 & 0.165 & 0.795 & 0.011 \\
    ARCTIC & 0.842 & 0.020 & 0.175 & 0.157 & 0.764 & 0.019 \\
    DexterHand & 0.998 & 0.019 & 0.256 & 0.155 & 0.782 & 0.013 \\
    Overall & 0.929 & 0.020 & 0.192 & 0.158 & 0.777 & 0.015 \\
    \bottomrule
  \end{tabular}}
\end{table}

\section{Reward Specification}
\label{sec:reward_details}
We specify the per-step reward as $r=r_g+r_s+r_p$ with task terms $r_g$, style terms $r_s$, and smoothness penalties $r_p$.
At reference frame $t$, let $q_t\in\mathbb{R}^D$ be the robot joint configuration and let $q_t^{\mathrm{ref}}$ be the reference target.
For each object $o$, let $p_{t,o}\in\mathbb{R}^3$ and $\bar{q}_{t,o}\in\mathbb{S}^3$ denote the simulated translation and unit quaternion, with targets $p^{\mathrm{ref}}_{t,o}$ and $\bar{q}^{\mathrm{ref}}_{t,o}$.
We partition $q_t$ into arm joints $q^a_t\in\mathbb{R}^{D_a}$ and finger joints $q^f_t\in\mathbb{R}^{D_f}$ and partition $q^{\mathrm{ref}}_t$ analogously into $q^{a,\mathrm{ref}}_t$ and $q^{f,\mathrm{ref}}_t$, with $D=D_a+D_f$.
We define tracking errors $e^a_t=q^a_t-q^{a,\mathrm{ref}}_t$, $e^f_t=q^f_t-q^{f,\mathrm{ref}}_t$, and $e^p_t=[p_{t,1}-p^{\mathrm{ref}}_{t,1};\dots;p_{t,O}-p^{\mathrm{ref}}_{t,O}]\in\mathbb{R}^{3O}$.
For object rotation we use the per-object quaternion angle error $\theta_{t,o}=2\arccos\left(\left|\bar{q}_{t,o}^\top \bar{q}^{\mathrm{ref}}_{t,o}\right|\right)$ and its mean $\bar{\theta}_t=\frac{1}{O}\sum_{o=1}^O \theta_{t,o}$.

Let $\dot q_t$ be joint velocities from the simulator and let $\ddot q_t$ be finite-difference joint accelerations using the simulator time step.
We define object linear velocities $\dot p_{t,o}$, angular velocities $\omega_{t,o}$, linear accelerations $\ddot p_{t,o}$, and angular accelerations $\dot\omega_{t,o}$ analogously, and stack them across objects into vectors $\dot p_t\in\mathbb{R}^{3O}$, $\omega_t\in\mathbb{R}^{3O}$, $\ddot p_t\in\mathbb{R}^{3O}$, and $\dot\omega_t\in\mathbb{R}^{3O}$.
For contact supervision, let $c_{t,o,\ell}\in\{0,1\}$ indicate a simulated contact between object $o$ and link $\ell$ at frame $t$, and let $y_{t,o,\ell}\in\mathbb{R}^3$ be the corresponding contact point in object coordinates, with targets $c^{\mathrm{ref}}_{t,o,\ell}$ and $y^{\mathrm{ref}}_{t,o,\ell}$.
We define $\mathcal{O}_{t,\ell}=\{o\mid c_{t,o,\ell}=1,\ c^{\mathrm{ref}}_{t,o,\ell}=1\}$, $\mathcal{L}_t=\{\ell\mid |\mathcal{O}_{t,\ell}|>0\}$, and $d_{t,\ell}^2=\frac{1}{|\mathcal{O}_{t,\ell}|}\sum_{o\in\mathcal{O}_{t,\ell}}\lVert y_{t,o,\ell}-y^{\mathrm{ref}}_{t,o,\ell}\rVert_2^2$.

For a vector $z\in\mathbb{R}^m$, we define the normalized squared error $\mathrm{mse}(z)=\frac{1}{m}\lVert z\rVert_2^2$.
Table~\ref{tab:reward_terms} lists the exact reward terms, kernel widths, and coefficients used in our experiments.

\begin{table}[t]
  \caption{Reward terms, functional forms, and coefficients used in our experiments. We use $\mathrm{mse}(z)=\lVert z\rVert_2^2/m$, where $m$ is the dimension of $z$.}
  \label{tab:reward_terms}
  \centering
  \scriptsize
  \setlength{\tabcolsep}{4pt}
  \begin{tabular}{@{}p{0.26\linewidth} p{0.54\linewidth} c@{}}
    \toprule
    Term & Expression & Weight \\
    \midrule
    \multicolumn{3}{@{}c}{\textit{Task Reward: $r_g$}} \\
    \midrule
    \texttt{tracking\_obj\_pos} & $\exp\!\left(-\mathrm{mse}(e^p_t)/\sigma_p^2\right)$, $\sigma_p=0.05$\,m & $+500$ \\
    \texttt{tracking\_obj\_rot} & $\exp\!\left(-\bar{\theta}_t^2/\sigma_\theta^2\right)$, $\sigma_\theta=0.5$\,rad & $+500$ \\
    \midrule
    \multicolumn{3}{@{}c}{\textit{Style Reward: $r_s$}} \\
    \midrule
    \texttt{tracking\_arm} & $\exp\!\left(-\mathrm{mse}(e^a_t)/\sigma_q^2\right)$, $\sigma_q=0.5$\,rad & $+10$ \\
    \texttt{tracking\_finger} & $\exp\!\left(-\mathrm{mse}(e^f_t)/\sigma_q^2\right)$, $\sigma_q=0.5$\,rad & $+10$ \\
    \texttt{contact\_reward} & $\frac{1}{L}\sum_{\ell=1}^L \mathbb{I}\left[\exists o,\ c_{t,o,\ell}=1,\ c^{\mathrm{ref}}_{t,o,\ell}=1\right]$ & $+100$ \\
    \texttt{contact\_distance\_}\allowbreak\texttt{reward} & $\frac{1}{|\mathcal{L}_t|}\sum_{\ell\in\mathcal{L}_t}\exp\!\left(-d_{t,\ell}^2/\sigma_c^2\right)$, $\sigma_c=0.03$\,m & $+100$ \\
    \midrule
    \multicolumn{3}{@{}c}{\textit{Penalty Term: $r_p$}} \\
    \midrule
    \texttt{qvel\_penalty\_arm} & $1-\exp\!\left(-\mathrm{mse}(\dot q^a_t)/\sigma_v^2\right)$, $\sigma_v=1.0$ & $-1$ \\
    \texttt{qvel\_penalty\_finger} & $1-\exp\!\left(-\mathrm{mse}(\dot q^f_t)/\sigma_v^2\right)$, $\sigma_v=1.0$ & $-1$ \\
    \texttt{qvel\_penalty\_obj\_pos} & $1-\exp\!\left(-\mathrm{mse}(\dot p_t)/\sigma_v^2\right)$, $\sigma_v=1.0$ & $-1$ \\
    \texttt{qvel\_penalty\_obj\_rot} & $1-\exp\!\left(-\mathrm{mse}(\omega_t)/\sigma_v^2\right)$, $\sigma_v=1.0$ & $-0.1$ \\
    \texttt{qacc\_penalty\_arm} & $1-\exp\!\left(-\mathrm{mse}(\ddot q^a_t)/\sigma_a^2\right)$, $\sigma_a=50.0$ & $-10$ \\
    \texttt{qacc\_penalty\_finger} & $1-\exp\!\left(-\mathrm{mse}(\ddot q^f_t)/\sigma_a^2\right)$, $\sigma_a=50.0$ & $-10$ \\
    \texttt{qacc\_penalty\_obj\_pos} & $1-\exp\!\left(-\mathrm{mse}(\ddot p_t)/\sigma_a^2\right)$, $\sigma_a=50.0$ & $-100$ \\
    \texttt{qacc\_penalty\_obj\_rot} & $1-\exp\!\left(-\mathrm{mse}(\dot\omega_t)/\sigma_a^2\right)$, $\sigma_a=50.0$ & $-10$ \\
    \bottomrule
  \end{tabular}
\end{table}

\section{Observation Specification}
\label{sec:obs_spec}
We use an asymmetric actor--critic. The actor observes a compact state that is sufficient for feedback tracking, while the critic receives additional signals that reduce value-estimation variance and make long-horizon training stable.

\paragraph{Actor observations.}
Let $q_t\in\mathbb{R}^D$ and $\dot q_t\in\mathbb{R}^D$ denote robot joint positions and velocities at reference index $t$. For each object $o\in\{1,\dots,O\}$, let $p_{t,o}\in\mathbb{R}^3$ and $\bar q_{t,o}\in\mathbb{S}^3$ denote object translation and unit quaternion, and let $\dot p_{t,o}\in\mathbb{R}^3$ and $\omega_{t,o}\in\mathbb{R}^3$ denote linear and angular velocities. Define the stacked pose vector $z_t=[q_t;\,p_{t,1};\,\bar q_{t,1};\dots;p_{t,O};\,\bar q_{t,O}]\in\mathbb{R}^{D+7O}$ and the stacked velocity vector $v_t=[\dot q_t;\,\dot p_{t,1};\,\omega_{t,1};\dots;\dot p_{t,O};\,\omega_{t,O}]\in\mathbb{R}^{D+6O}$. The actor input is
\begin{equation}
  o^{\mathrm{actor}}_t=[z_{t-d};\,v_{t-d}]\in\mathbb{R}^{2D+13O},
\end{equation}
where $d\ge 0$ is an environment-step observation delay sampled during training. We add zero-mean Gaussian noise to the actor observation terms during training.

\paragraph{Critic observations.}
The critic receives the current uncorrupted pose and velocity vectors $z_t$ and $v_t$. It also receives binary contact indicators $c_t\in\{0,1\}^{LO}$ that mark whether each of $L$ tracked hand links is in contact with each object at index $t$. In addition, the critic observes a short stack of future reference poses. Let $K$ be the number of future reference indices. Define $z^{\mathrm{ref}}_t\in\mathbb{R}^{D+7O}$ analogously to $z_t$ using the reference targets and stack it as $z^{\mathrm{ref}}_{t:t+K}\in\mathbb{R}^{(K+1)(D+7O)}$. The critic also receives reference contact flags $c^{\mathrm{ref}}_t\in\{0,1\}^{LO}$ and a normalized phase scalar $\varphi_t=t/(T-1)\in[0,1]$. The critic input concatenates these terms:
\begin{equation}
  o^{\mathrm{critic}}_t=[z_t;\,v_t;\,c_t;\,z^{\mathrm{ref}}_{t:t+K};\,c^{\mathrm{ref}}_t;\,\varphi_t].
\end{equation}

\section{Controller and Reset Hyperparameters}
\label{sec:hyperparams}
ConTrack uses one scalar controller for the task--style balance and one reset distribution for long-horizon stability. Unless stated otherwise, all hyperparameters below are held fixed across clips and tiers.

\paragraph{Dual controller.}
We use target ratio $\alpha=0.9$ in the main results, step size $\eta=0.05$ in Eq.~\ref{eq:dual_update}, and initialize $\lambda=0$ so that $w_{\mathrm{task}}=\sigma(\lambda)=0.5$ at the start of training. We compute $\hat{J}_g$ as the mean episodic task return over the most recent $100$ completed episodes and set $J_g^\star$ as the maximum value of $\hat{J}_g$ over the most recent $20$ updates. In practice, $\alpha$ is the main user-facing knob. Decreasing $\alpha$ allocates more optimization capacity to style, and Fig.~\ref{fig:tradeoff_alpha_sweep} sweeps $\alpha$ to trace a trade-off curve.

\paragraph{Reset library.}
We update the per-frame continuation length statistic $\bar{\ell}_k$ with an exponential moving average using coefficient $0.05$. We sample reset frames from $p(k)\propto \exp(-u_k/\tau)$ with temperature $\tau=1.0$. With probability $0.5$, we instead sample a failure-boundary frame from the top $64$ sharp changes in $u_k$ and shift the reset $100$ frames earlier.

\section{Contact Annotation Processing}
\label{sec:contact_processing}
Reference clips store contact labels and points at the capture segment level. For each object, we store a binary contact tensor and object-local contact points for 15 hand segments. Contact is extracted offline with a fixed proximity threshold of $5$\,mm and stored in the reference file.

ConTrack uses a link-level contact prior that matches the xHand embodiment. We map the 15 segments into 10 link groups, and we aggregate segment-level contacts into per-link binary events and contact points. For each link group $\ell$ with segment index set $\mathcal{S}_\ell$, we define
\begin{equation}
  c^{\mathrm{ref}}_{t,o,\ell}=\max_{s\in\mathcal{S}_\ell} c^{\mathrm{seg}}_{t,o,s},\qquad
  y^{\mathrm{ref}}_{t,o,\ell}=\frac{\sum_{s\in\mathcal{S}_\ell} c^{\mathrm{seg}}_{t,o,s}\,y^{\mathrm{seg}}_{t,o,s}}{\max\left(\sum_{s\in\mathcal{S}_\ell} c^{\mathrm{seg}}_{t,o,s},1\right)},
\end{equation}
where $c^{\mathrm{seg}}_{t,o,s}\in\{0,1\}$ and $y^{\mathrm{seg}}_{t,o,s}\in\mathbb{R}^3$ are the segment-level contact event and object-local point. Missing points are stored as NaN in the reference and treated as zero, so the aggregation reduces to a well-defined average on contacting segments.

The segment-to-link grouping is
\begin{equation}
  \begin{aligned}
    \texttt{index\_link1}:  & \ \{\texttt{index\_proximal}\},                        \\
    \texttt{index\_link2}:  & \ \{\texttt{index\_middle},\texttt{index\_distal}\},   \\
    \texttt{middle\_link1}: & \ \{\texttt{middle\_proximal}\},                       \\
    \texttt{middle\_link2}: & \ \{\texttt{middle\_middle},\texttt{middle\_distal}\}, \\
    \texttt{ring\_link1}:   & \ \{\texttt{ring\_proximal}\},                         \\
    \texttt{ring\_link2}:   & \ \{\texttt{ring\_middle},\texttt{ring\_distal}\},     \\
    \texttt{pinky\_link1}:  & \ \{\texttt{pinky\_proximal}\},                        \\
    \texttt{pinky\_link2}:  & \ \{\texttt{pinky\_middle},\texttt{pinky\_distal}\},   \\
    \texttt{thumb\_link1}:  & \ \{\texttt{thumb\_proximal}\},                        \\
    \texttt{thumb\_link2}:  & \ \{\texttt{thumb\_middle},\texttt{thumb\_distal}\}.
  \end{aligned}
\end{equation}
We align these targets to the action step rate using nearest-neighbor sampling in time.

\section{Interaction Budget and Compute}
\label{sec:compute}

All experiments use a single NVIDIA GeForce RTX 4090 GPU, we train ConTrack, ManipTrans, and DexMachina for $5000$ PPO updates per clip. For our method, training for $5000$ updates takes about $1.9$ to $2.9$ hours per clip. Each update collects $16$ action steps per environment, and we select the number of parallel environments per clip to saturate GPU throughput. In our benchmark runs, this yields roughly $5.6\times 10^8$ to $6.4\times 10^8$ simulator steps per clip. Since the rollout collection size per update is fixed, mid-trajectory resets change the distribution of visited reference indices but not the total interaction budget.

\section{DexterHand Ring Failure Analysis}
\label{sec:appendix_dexterhand_ring_failure}
DexterHand Ring is the least reliable clip under the fixed $5000$-update protocol. To separate feasibility from budget, we train ConTrack longer on this clip. With $100\,000$ PPO updates, ConTrack reaches a $94\%$ success rate under the same pose-break thresholds. We exclude this extended-budget result from the fixed-budget tables to keep the comparison protocol consistent.

To localize the failure mode at $5000$ updates, we analyze $4096$ evaluation episodes under the same termination rule. We find that $82\%$ of terminations are triggered by the rotation pose-break condition. At termination, the mean object translation error remains $2.2$\,cm, which indicates that failures are driven primarily by loss of orientation rather than gross positional drift. Termination indices also concentrate around action step $81$ out of the $T=297$ action steps of the clip, suggesting a phase-specific instability concentrated in a rotation-dominant segment.

Taken together, these statistics point to a narrow contact--rotation transition as the bottleneck. More accurate modeling of rotational contact dynamics and training curricula that align reset timing to fast rotational phases are promising directions for extending ConTrack in this regime.

\end{document}